\documentclass{article}

\usepackage{microtype}
\usepackage{graphicx}
\usepackage{booktabs} 
\usepackage{multirow}
\usepackage{amssymb}
\usepackage{amsmath}
\usepackage{subcaption}
\usepackage{bm}
\usepackage{xspace}
\usepackage{enumitem}
\usepackage[table,xcdraw]{xcolor}
\usepackage{wrapfig}

\usepackage{hyperref}

\usepackage[accepted]{icml2020}

\icmltitlerunning{Meta-Learning with Shared Amortized Variational Inference}

\def\Eq#1{Eq.~(\ref{eq:#1})}

\def\be{\begin{equation}}
\def\ee{\end{equation}}
\def\bea{\begin{eqnarray}}
\def\eea{\end{eqnarray}}
\def\fig#1{Figure~\ref{fig:#1}}
\def\tab#1{Table~\ref{tab:#1}}
\def\sect#1{Section~\ref{sec:#1}}

\def\R{{\rm I\!R}}

\def\diag#1{\textrm{diag}\left( #1 \right)}

\makeatletter
\DeclareRobustCommand\onedot{\futurelet\@let@token\@onedot}
\def\@onedot{\ifx\@let@token.\else.\null\fi\xspace}
\def\eg{{e.g}\onedot, }

\def\ie{{i.e}\onedot, }

\newcommand{\vect}[1]{\boldsymbol{#1}}

\def\mypar#1{\vspace{0mm}{\noindent\bf #1.}\hspace{1mm}}

\newcommand{\myskip}[1]{}

\begin{document}

\twocolumn[
\icmltitle{Meta-Learning with Shared Amortized Variational Inference}

\icmlsetsymbol{equal}{*}

\begin{icmlauthorlist}
\icmlauthor{Ekaterina Iakovleva}{ia}
\icmlauthor{Jakob Verbeek}{fb}
\icmlauthor{Karteek Alahari}{ia}
\end{icmlauthorlist}

\icmlaffiliation{ia}{Univ.\ Grenoble Alpes,\ Inria,\ CNRS,\ Grenoble INP,\ LJK,\ 38000 Grenoble,\ France.}
\icmlaffiliation{fb}{Facebook Artificial Intelligence Research,  Work done while Jakob Verbeek was at Inria}

\icmlcorrespondingauthor{Ekaterina Iakovleva}{ekaterina.iakovleva@inria.fr}

\icmlkeywords{Meta Learning, Bayesian inference, Bayesian Meta Learning, Conditional Autoencoder, ICML}

\vskip 0.3in
]

\printAffiliationsAndNotice{}

\begin{abstract}
We propose a novel amortized variational inference scheme for an empirical Bayes meta-learning model, where model parameters are treated as latent variables.
We learn the  prior distribution over model parameters conditioned on limited training data using a variational autoencoder approach.
Our framework proposes sharing the same amortized inference network between the conditional prior and variational posterior distributions over the model parameters. 
While the posterior leverages both the labeled support and query data, the conditional prior is based only on the labeled support data.
We show that in earlier work, relying on Monte-Carlo approximation, the conditional prior collapses to a Dirac delta function. 
In contrast, our variational approach prevents this collapse and preserves uncertainty over the model parameters. 
We evaluate our approach on the miniImageNet, CIFAR-FS 
and FC100 datasets, and present results demonstrating its advantages over previous work.
\end{abstract}

\section{Introduction}
\label{sec:introduction}

\begin{figure*}[t]
\begin{center}
\scalebox{.9}{\includegraphics[width=\textwidth]{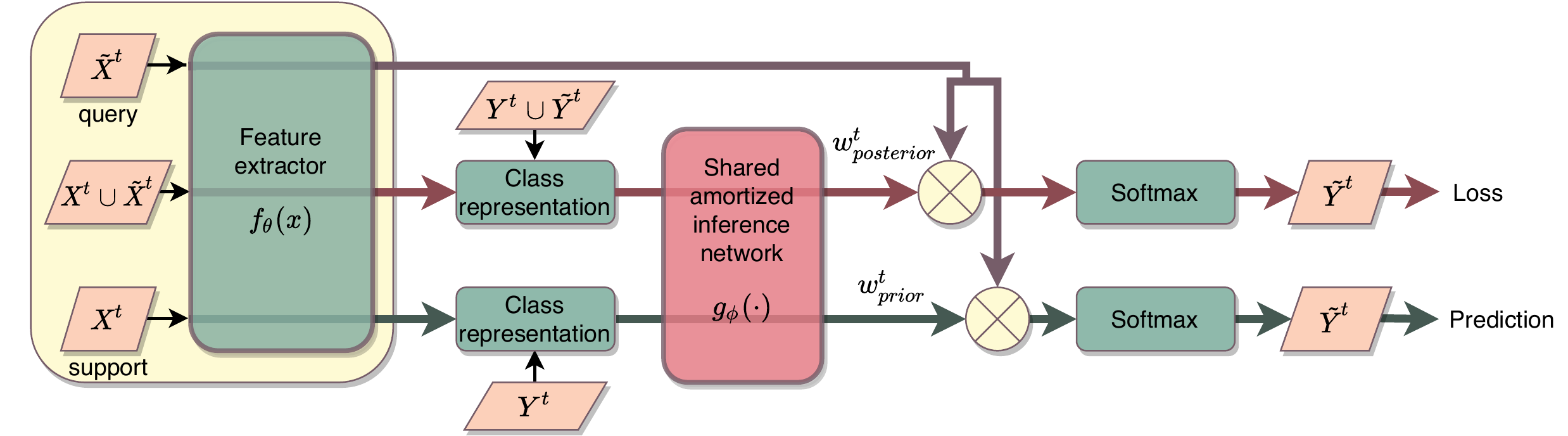}}
\end{center}
\caption{SAMOVAR, our meta-learning model for few-shot image classification. For task $t$, query data $\tilde{X}^t$ and support data $X^t$ are put through a task-agnostic feature extractor $f_{\theta}(x)$. 
The features are then averaged class-wise, and mapped by the shared amortized inference network into prior and posterior over the task-specific classifier weight vectors. 
Classifiers $w^t_{posterior}$ and $w^t_{prior}$ sampled from these distributions map query features $f_{\theta}(\tilde{X}^t)$ to predictions on the query labels $\tilde{Y}^t$ used in training and testing, respectively.}
\label{fig:SAMOVAR}
\end{figure*}

While  people have an outstanding ability to learn from just a few examples,  generalization from small sample sizes has been one of the  long-standing goals of  machine learning. 
Meta-learning, or ``learning to learn'' \cite{schmidhuber99meta}, aims to improve generalization in small sample-size settings by leveraging the experience of having learned to solve  related tasks in the past.
The core idea is to learn a  meta model that, for any given task,  maps a small set of training samples for a new task to a model that  generalizes well.
 
A recent surge of interest in meta-learning has explored a wide spectrum of approaches.
This includes nearest neighbor based methods \cite{guillaumin09iccv2,vinyals2016matching}, nearest class-mean approaches \cite{dvornik19iccv,mensink12eccv,ren18fewshotssl, snell2017prototypical}, 
 optimization based methods 
\cite{finn17icml,ravi17iclr}, adversarial approaches \cite{zhang2018metagan}, and  Bayesian models \cite{gordon2018metalearning,grant2018recasting}.
The  Bayesian approach is particularly interesting, since it provides a coherent framework to reason about model uncertainty, not only in small sample-size settings, but also others such as incremental learning \cite{kochurov18iclr}, and  ensemble learning \cite{gal16icml}.
Despite its attractive properties, intractable integrals over  model parameters or other latent variables,  which are at the heart of the Bayesian framework, make it often necessary to turn to stochastic Monte Carlo or analytic approximations for  practical  implementations.

In our work, we follow the Bayesian latent variable approach, and learn a  prior on the parameters of the classification model conditioned on a  small training sample set for the task.
We use a variational inference framework to approximate the intractable marginal likelihood function during training. 
The variational distribution approximates the posterior on the  parameters of the classification model, given training and test data.
Both the prior and posterior are parameterized as deep neural networks that take a set of labeled data points as input. 
By sharing the inference network across these two distributions, we leverage more data to learn these conditionals and avoid overfitting.
\fig{SAMOVAR} illustrates the overall structure of our model, SAMOVAR.

We compare the variational training approach with the Monte Carlo approach followed by \citet{gordon2018metalearning} on synthetic data.
We find that when using a small number of samples for stochastic back-propagation in the Monte Carlo approach, which results in faster training, the prior collapses to a Dirac delta, and the model degenerates to a deterministic parameter generating network.
In contrast, our variational training approach does not suffer from this deficiency, and leads to an accurate estimation of the  variance.
Experiments on few-shot image classification using the miniImageNet, CIFAR-FS and FC100 datasets confirm these findings, and we observe improved accuracy using the variational approach to train the VERSA model~\cite{gordon2018metalearning}. 
Moreover, we use the same variational framework to train a stochastic version of the TADAM few-shot image classification model \cite{oreshkin18nips}, replacing the deterministic prototype classifier with a scaled cosine classifier with stochastic weights.
Our stochastic formulation significantly improves performance over the base architecture, and yields results competitive with the state of the art on the miniImageNet, CIFAR-FS and FC100 datasets.

\section{Related Work}

\mypar{Distance-based classifiers}
A straightforward approach to handle small training sets is to use nearest neighbor \cite{weinberger06nips,guillaumin09iccv2,vinyals2016matching}, or nearest prototype \cite{mensink12eccv, snell2017prototypical,dvornik19iccv,ren18fewshotssl,oreshkin18nips} classification methods.
In a ``meta'' training phase, a metric -- or, more generally, a data representation -- is learned using samples from a large number of  classes. 
At test time, the learned metric can then be used to classify samples across a set of classes not seen during training, by relying on distances to individual samples or ``prototypes,'' \ie per-class averages.
Alternatively, it is also possible to learn a network that takes two samples as input and predicts whether they belong to the same class \cite{sung2018learning}. Other work has explored the use of task-adaptive metrics, by conditioning the feature extractor on the class prototypes for the task at hand \cite{oreshkin18nips}.
We show that our latent variable approach is complementary and improves the effectiveness of the latter task conditioning scheme.

\mypar{Optimization-based approaches}
Deep neural networks are typically learned from large datasets using SGD. 
To adapt to the regime of (very) small training datasets, optimization-based meta-learning techniques replace the vanilla SGD approach by a trainable update mechanism \cite{bertinetto2018metalearning,finn17icml,ravi17iclr}, \eg 
by learning a parameter initialization, such that a small number of SGD updates yields good performance \cite{finn17icml}.
In addition to parameter initialization,
the use of an LSTM model to control the influence of the gradient for updating the current parameters has also been explored~
\cite{ravi17iclr}.
In our work, the amortized inference network makes a single feed-forward pass through data to estimate a distribution on the parameters, instead of multiple passes to update the parameters.

\mypar{Latent variable models}
Gradient-based estimators of the parameters have a high variance in the case of small sample sizes. 
It is natural to explicitly model this variance by treating the parameters as latent variables in a Bayesian framework \cite{garnelo18icmlw,gordon2018metalearning,grant2018recasting,kim19iclr,mackay92phd,neal95phd}.
The marginal likelihood of the test labels given the training set is then obtained by integrating out the latent model parameters.
This typically intractable marginal likelihood, required for training and prediction, can be approximated using (amortized) variational inference \cite{garnelo18icmlw,kim19iclr}, Monte Carlo sampling \cite{gordon2018metalearning}, or a Laplace approximation \cite{grant2018recasting}.
Neural processes \cite{garnelo18icmlw,kim19iclr} are also related to our work in their structure, and the use of shared inference network between the prior and variational posterior. 
Where neural processes use the task-specific latent variable as an additional input to the classifier network, we explicitly model the parameters of a linear classifier as the latent variable. This increases interpretability of the latent space, and allows for a flexible number of classes.

Interestingly, some optimization-based approaches can be viewed as approximate inference methods in latent variable models \cite{grant2018recasting,rusu19iclr}.
Semi-amortized inference techniques \cite{marino18icml,kim18icml},  which combine feed-forward parameter initialization and iterative gradient-based refinement of the approximate posterior, can be seen as a hybrid of optimization-based and Bayesian approaches. Deterministic approaches that generate a single parameter vector for the task model, given a set of training samples  \cite{bertinetto16nips,ha17iclr,qiao2018few}, can be seen as a special case of the latent variable model with Dirac delta conditional distributions on the parameters.

\section{Our Meta-Learning Approach}
\label{sec:model_description}

We follow the common meta-learning setting of episodic training of $K$-shot $N$-way classification on the {\it meta-train} set with $C$ classes  \citep{finn17icml,gordon2018metalearning,ravi17iclr}.
For each classification task $t$ sampled from a distribution over tasks $p(\mathcal{T})$, the training data $D^t = \{(\vect{x}_{k,n}^t, \vect{y}_{k,n}^t)\}_{k,n=1}^{K,N}$ ({\it support} set) consists of $K$ pairs of samples $\vect{x}_{k,n}^t$ and their  labels $\vect{y}_{k,n}^t$ from each of $N$ classes. 
The meta-learner takes the $KN$ labeled samples as input, and outputs a classifier across these $N$ classes to classify $MN$ unlabeled samples from the testing data $\tilde{D}^t = \{(\vect{\tilde{x}}_{m,n}^t, \vect{\tilde{y}}_{m,n}^t)\}_{m,n=1}^{M,N}$ ({\it query} set). 
During the {\it meta-train} stage, the meta-learner iterates over $T$ episodes where each episode corresponds to a particular task $t$. 
During the {\it meta-test} stage, the model is presented with new tasks where the support and query sets are sampled from the meta-test set, which consists of previously unseen classes $C'$. 
The support set is used as input to the trained meta-learner, and the classifier produced by  meta-learning is used to evaluate the performance on the query set. 
Results are averaged over a large set of meta-test tasks.

In this section, we propose a probabilistic framework for meta-learning. In \sect{generative_model}, we start with a description of the multi-task graphical model that we adopt. We then derive an amortized variational inference with learnable prior for this generative model in \sect{shared_cvae}, and propose to share the amortized networks for prior and approximate posterior. Finally, in \sect{samovar_design} we describe the design of our model, SAMOVAR, which is trained with the proposed shared variational inference method.
    
\subsection{Generative Meta-Learning Model}
\label{sec:generative_model}

\begin{figure}
\begin{center}
\scalebox{.7}{\includegraphics[width=\columnwidth]{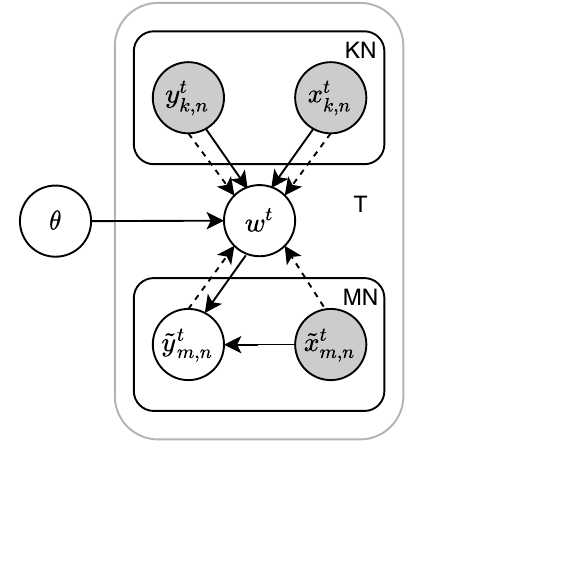}}
\end{center}
\vspace{-3.5\baselineskip}
\caption{Hierarchical graphical model. The solid lines correspond to the generative process, while the dashed lines correspond to the variational inference procedure.
Shaded nodes represent observed variables, non-shaded ones correspond to latent variables.
}
\label{fig:directed_graphical_model}
\end{figure}

We employ a hierarchical graphical model shown in \fig{directed_graphical_model}.
This multi-task model includes latent parameters $\theta$, shared across all the $T$ tasks, and task-specific latent parameters  $\{w^t\}_{t=1}^T$. 
The marginal likelihood of the query labels $\tilde{Y}=\{\tilde{Y}^t\}_{t=1}^T$, given the query samples $\tilde{X}=\{\tilde{X}^t\}_{t=1}^T$ and the support sets $D = \{D^t\}_{t=1}^T$, is obtained as
\begin{equation}
\label{eq:marginal_likelihood}
    \begin{split}
    p(&\tilde{Y}|\tilde{X},D) = \\
    &\int p(\theta) 
    \prod_{t=1}^T \int 
    p(\tilde{Y}^t|\tilde{X}^t,w^t) 
    p(w^t|D^t, \theta)
    \textrm{d}w^t\textrm{d}\theta.
    \end{split}
\end{equation}
The first term, $p(\theta)$, is the prior over the global task-independent parameters $\theta$.
The second term, $p(\tilde{Y}^t|\tilde{X}^t,w^t)$, is the likelihood of query labels $\tilde{Y}^t$, given query samples $\tilde{X}^t$ and task-specific  parameters $w^t$. 
For example, this could be a linear classifier with weights $w^t$ over features computed by a network with parameters $\theta$.
The third term, $p(w^t|D^t, \theta)$ is the conditional distribution on the task parameters $w^t$ given the 
support set $D^t$ and global parameters $\theta$. 
We parameterize this distribution with a deep neural network with parameters $\phi$ as
$p_{\phi}\left(w^t|D^t, \theta\right)$. 

Following \citet{gordon2018metalearning,grant2018recasting,hue2020empirical}, we consider a point estimate for $\theta$ to simplify the model. The per-task marginal likelihood is then
\bea
 \hspace{-0.9em}p(\tilde{Y}^t|\tilde{X}^t,D^t, \theta)\hspace{-0.9em}& = &\hspace{-0.9em}\int p(\tilde{Y}^t|\tilde{X}^t,w^t)p_{\phi}(w^t|D^t, \theta)\textrm{d}w^t, \label{eq:distribution_over_outputs}
\\
 \hspace{-0.9em}p(\tilde{Y}^t|\tilde{X}^t,w^t)\hspace{-0.9em}& = &\hspace{-0.9em}\prod_{m=1}^M p(\tilde{\vect{y}}_m^t|\tilde{\vect{x}}_m^t,w^t).
\eea

To train the model, a  Monte Carlo approximation  of the integral in \Eq{distribution_over_outputs} was used in  \citet{gordon2018metalearning}:
\begin{equation}
    \mathcal{L}(\theta,\phi)=\frac{1}{TM}\sum_{t=1}^{T}\sum_{m=1}^M\log\frac{1}{L}\sum_{l=1}^L p(\tilde{\vect{y}}_m^t|\tilde{\vect{x}}_m^t,w_l^t),
    \label{eq:MonteCarlo}
\end{equation}
where $w_l^t\sim p_{\phi}(w^t|D^t, \theta)$.
In our experiments in \sect{experiments}, we show that  training with this approximation tends to severely underestimate the variance in $p_{\phi}(w^t|D^t, \theta)$,  effectively reducing the model to a deterministic one, and defying the use of a stochastic latent variable model.

\subsection{Shared Amortized Variational Inference}
\label{sec:shared_cvae}

To prevent the conditional prior $p_{\phi}(w^t|D^t, \theta)$ from degenerating, we use amortized variational inference \cite{kingma14iclr,rezende14icml} to approximate the intractable true posterior $p(w^t|\tilde{Y}^t,\tilde{X}^t,D^t, \theta)$.
Using the  approximate posterior $q_{\psi}(w^t|\tilde{Y}^t,\tilde{X}^t,D^t,\theta)$ 
parameterized by $\psi$, we obtain  the variational evidence lower bound (ELBO) of \Eq{distribution_over_outputs} as
\begin{equation}
    \begin{split}
    \log &p(\tilde{Y}^t|\tilde{X}^t,D^t, \theta) \geq \mathbb{E}_{q_{\psi}}\left[\log p(\tilde{Y}^t|\tilde{X}^t, w^t)\right] \\ 
    &-\mathcal{D}_\textrm{KL}\left(q_{\psi}(w^t|\tilde{Y}^t,\tilde{X}^t, D^t, \theta) || p_{\phi}(w^t|D^t, \theta)\right).
    \end{split}
    \label{eq:ELBO}
\end{equation}
The first term can be interpreted as a reconstruction loss, that reconstructs the labels of the query set using latent variables $w^t$ sampled from the approximate posterior, and the second term as a regularizer that encourages the approximate posterior to remain close to the conditional prior $p_{\phi}(w^t|D^t, \theta)$. 
We approximate the reconstruction term using $L$ Monte Carlo samples, and add a regularization coefficient $\beta$ to weigh the KL term \cite{higgins2017beta}. 
With this, our optimization objective is:
\begin{equation}
    \begin{split}
    &\hat{\mathcal{L}}(\Theta)
    =\frac{1}{T}\sum_{t=1}^{T}\left[\sum_{m=1}^M\frac{1}{L}\sum_{l=1}^L\log p(\tilde{\vect{y}}_m^t|\tilde{\vect{x}}_m^t,w_l^t)\right. \\
    &\left.\vphantom{\sum_{l=1}^L} -\beta\mathcal{D}_\textrm{KL}\left(q_{\psi}(w^t|\tilde{Y}^t,\tilde{X}^t,D^t, \theta) || p_{\phi}(w^t|D^t, \theta)\right)\right],
    \end{split}
\end{equation}
where $w_l^t\sim q_{\psi}(w|\tilde{Y}^t,\tilde{X}^t, D^t, \theta)$. 
We maximize the ELBO w.r.t. $\Theta=\{\theta, \phi, \psi\}$ to jointly train the model parameters $\theta$, $\phi$, and the variational parameters $\psi$.

We use Monte Carlo sampling from the learned model to make predictions at test time as: 
\begin{equation}
    \label{eq:predictive_distribution}
    p(\tilde{\vect{y}}^t_m|\tilde{\vect{x}}^t_m,D^t, \theta)
    \approx  \frac{1}{L}\sum_{l=1}^L  p(\tilde{\vect{y}}^t_m|\tilde{\vect{x}}^t_m,w_l^t),
\end{equation}
where $w_l^t\sim p_{\phi}(w^t|D^t, \theta)$. 
In this manner, we leverage the stochasticity of our model by averaging predictions over multiple realizations of $w^t$.

The approach presented above suggests to train  separate networks to parameterize the conditional prior $p_{\phi}(w^t|D^t,\theta)$ and the approximate posterior $q_{\psi}(w^t|\tilde{Y}^t,\tilde{X}^t, D^t,\theta)$. 
Since in both cases the conditioning data consists of labeled samples, it is possible to share the network for both distributions, and simply change the input of the network to obtain one distribution or the other. 
Sharing has two  advantages: (i) It reduces the number of parameters to train, decreasing the memory footprint of the model and the risk of overfitting. (ii) It facilitates the learning of a non-degenerate prior.

Let us elaborate on the second point.
Omitting all dependencies for brevity,
the KL divergence 
$
    \mathcal{D}_\textrm{KL}(q||p)=\int q(w)\left[\log q(w)- \log p(w) \right]
$
in \Eq{ELBO} compares the posterior $q(w)$ and the prior $p(w)$.
Consider the case when the prior converges to a Dirac delta,  while the posterior does not. 
Then, there exist points in the support of the posterior for which $p(w) \approx 0$, therefore, the KL divergence tends to infinity.
The only alternative in this case is for the posterior to converge to the same Dirac delta. 
This would mean that for different inputs the inference network produces the same (degenerate) distribution. In particular, the additional conditioning data available in the posterior would leave the distribution  unchanged, failing to learn from the additional data.
While in theory this is possible, we do not observe it in practice. 

We coin our approach  ``SAMOVAR'',  short for Shared AMOrtized VARiational inference.

\subsection{Implementing SAMOVAR: Architectural Designs}
\label{sec:samovar_design}

The key properties we expect SAMOVAR to have are: (i) the ability to perform the inference in a feed-forward way (unlike gradient-based models), and (ii) the ability to handle a variable  number of classes  within the tasks. 
We build upon the work of \citet{gordon2018metalearning,qiao2018few}, to meet both these requirements.
We start with VERSA \cite{gordon2018metalearning} where the feature extractor is followed by an amortized inference network, which returns a linear classifier with stochastic weights.
SAMOVAR-base, our baseline architecture built this way on VERSA, consists of the following components.

\mypar{Task-independent feature extractor} 
We use a deep convolutional neural network (CNN),  $f_{\theta}$, shared across all tasks,  to embed  input images  $\vect{x}$ in $\R^d$.
The extracted features are the only information from the samples used in the rest of the model.
The CNN architectures used for different datasets are detailed in \sect{samovar_implementation}.

\mypar{Task-specific linear classifier} 
Given the  features, we use multi-class logistic discriminant classifier, with task-specific weight matrix $w^t\in\R^{N\times d}$.
That is, for the query samples $\vect{\tilde{x}}$ we obtain a distribution over the labels as:
\begin{equation}
p(\tilde{\vect{y}}_m^t|\tilde{\vect{x}}_m^t,w^t) = 
   \text{softmax}\left(w^t f_{\theta}(\vect{\tilde{x}}_m^t)\right).
   \label{eq:classif}
\end{equation}
    
\mypar{Shared amortized inference network} 
{We use a deep permutation invariant network $g_{\phi}$ to parameterize the prior over the task-specific weight matrix $w^t$, given a set of  labeled samples.
The distribution on $w^t$ is factorized over its rows $w_1^t, \dots, w_N^t$ to allow for variable number of classes, and to simplify the structure of the model.
For any class $n$, the inference network $g_{\phi}$ maps the corresponding set of support feature embeddings  $\{f_{\theta}(\vect{x}_{k,n}^t)\}_{k=1}^K$ to the parameters of a distribution over $w_n^t$.}
We use a Gaussian with diagonal covariance to model these distributions on the weight vectors, \ie
\bea
p_{\phi}(w^t_n|D^t,\theta) =  \mathcal{N}(\vect{\mu}_n^t,\diag{\vect{\sigma}_n^t}),
\eea
where the mean and the variance are computed by the inference network as:
\bea
  \begin{bmatrix} \vect{\mu}_n^t \\ \vect{\sigma}_n^t\end{bmatrix} = g_{\phi}\left(\frac{1}{K}\sum_{k=1}^{K}f_{\theta}(\vect{x}_{k,n}^t)\right).
\eea
{To achieve permutation invariance among the samples, we average the feature vectors within each class before feeding them into the inference network $g_{\phi}$. The approximate variational posterior is obtained in the same manner, but in this case the feature average that is used as input to the inference network is computed over the union of labeled support and query samples.}

To further improve the model, we employ techniques commonly used in meta-learning classification models: scaled cosine similarity, task conditioning, and auxiliary co-training.

\mypar{Scaled cosine similarity} 
Cosine similarity based classifiers have recently been  widely adopted in few-shot classification \cite{dvornik19iccv,gidaris2019boosting,lee2019meta,oreshkin18nips,ye2018learning}. 
Here, the  linear classifier is replaced with a  classifier based on the cosine similarity with the  weight vectors $w_n^t$, scaled with a temperature parameter $\alpha$:
\begin{equation}
\label{eq:reconstruction_loss}
    p(\vect{\tilde{y}}_m^t|\vect{\tilde{x}}_m^t,w_n^t) =
    \text{softmax}\left( \alpha\frac{f_{\theta}(\vect{\tilde{x}}_m^t)^\top w_n^t}{||f_{\theta}(\vect{\tilde{x}}_m^t)||\cdot ||w_n^{t}||} \right)
\end{equation}
We refer this version of our model as SAMOVAR-SC.

\mypar{Task conditioning} 
A limitation of the above models is that the  weight vectors $\vect{w}_n^t$ depend only on the samples of class $n$. 
To leverage the full context of the task, we adopt the  task embedding network (TEN) of \citet{oreshkin18nips}. 
For each feature dimension of $f_{\theta}$, TEN provides an affine transformation conditioned on the task data, similar to FiLM conditioning layers \cite{perez2018film} and conditional batch normalization \cite{pmlr-v80-munkhdalai18a,dumoulin17iclr2}.
In particular, input to TEN is the average $\vect{c} = \frac{1}{N}\sum_n \vect{c}_n$, of the per-class prototypes, $\vect{c}_n = \frac{1}{K} \sum_k f_\theta(\vect{x}_{kn}^t)$  in the task $t$, and outputs are translation and scale parameters for all feature channels in the feature extractor layers. 
In SAMOVAR, we use TEN to modify both the support and query features $f_{\theta}$ before they enter the inference network $g_\phi$. The query features that enter into the linear/cosine classifiers are left unchanged.

\mypar{Auxiliary co-training} 
Large feature extractors can benefit from 
auxiliary co-training to prevent overfitting, stabilize the training, and boost the performance \cite{oreshkin18nips}.
{We leverage this by sharing the feature extractor $f_{\theta}$ of the meta-learner with an auxiliary classification task across all the classes in the  meta-train set, using the cross-entropy loss for a linear logistic classifier over $f_{\theta}$.}

\begin{figure*}[t!]
    \centering
    \begin{subfigure}[t]{0.32\textwidth}
    \includegraphics[width=\textwidth]{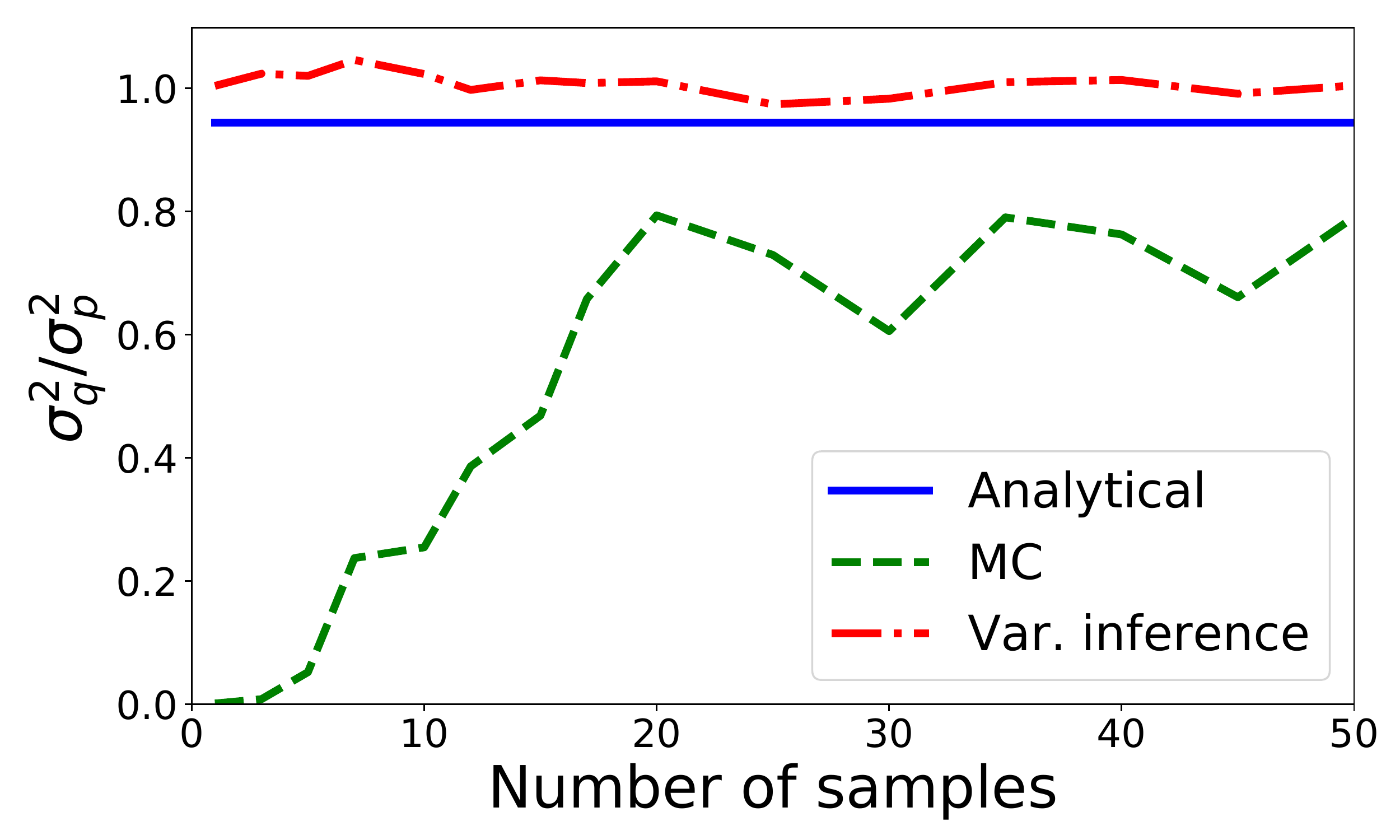}
    \caption{$\sigma_y =0.1$}
    \label{fig:synthetic_relative_variance_01}
    \end{subfigure}
    \begin{subfigure}[t]{0.32\textwidth}
    \includegraphics[width=\textwidth]{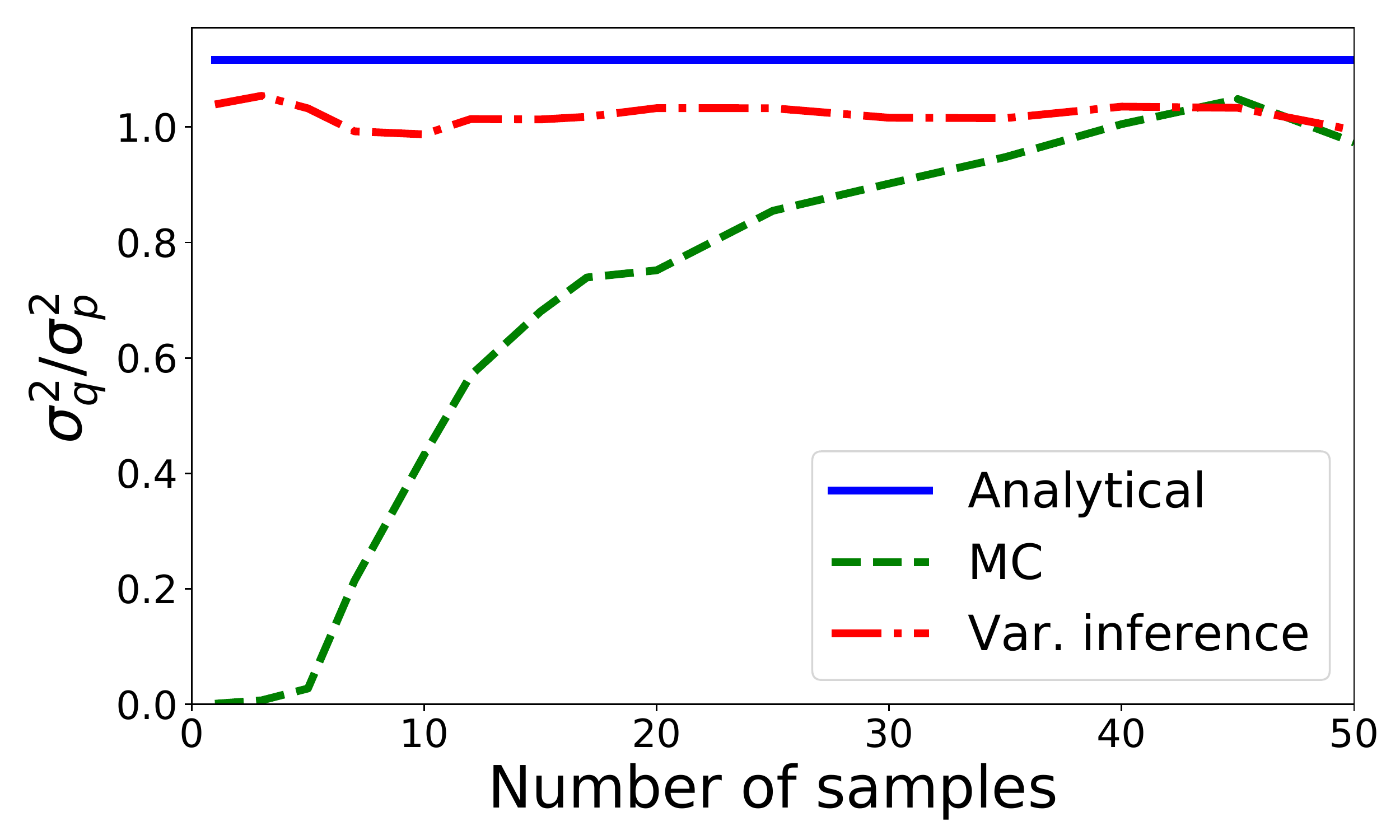}
    \caption{$\sigma_y =0.5$}
    \label{fig:synthetic_relative_variance_05}
    \end{subfigure}
    \begin{subfigure}[t]{0.32\textwidth}
    \includegraphics[width=\textwidth]{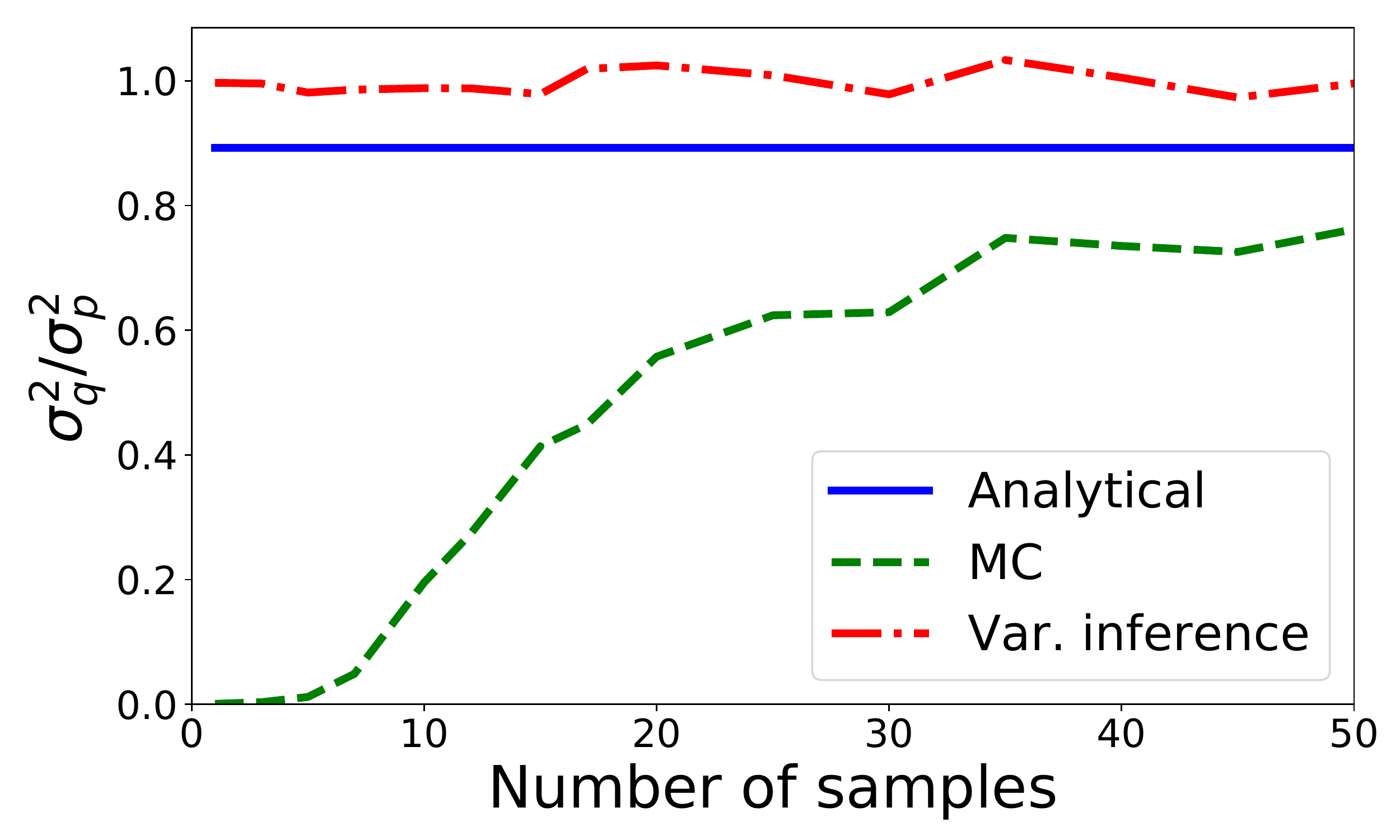}
    \caption{$\sigma_y =1.0$}
    \label{fig:synthetic_relative_variance_10}
    \end{subfigure}
    \caption{Ratio between the variance in $\psi$ estimated by the trained inference network $q_\phi(\psi|D^t)$ and $\sigma^2_p$ in true posterior $p(\psi|D^t)$, for different number of samples $L$ from the inference network during training.
    }
    \label{fig:synthetic_relative_sigma}
\end{figure*}

\section{Experiments}
\label{sec:experiments}

We analyze  the differences between training with Monte Carlo estimation and variational inference with a controlled synthetic data experiment in \sect{synthetic_task}.
Then, we present the few-shot image classification experimental setup in \sect{samovar_implementation}, followed by results, and a comparison to related work in \sect{results}.

\subsection{Synthetic Data Experiments}
\label{sec:synthetic_task}

We consider the same hierarchical generative process as  \citet{gordon2018metalearning}, which allows for exact inference:
\begin{equation}
p(\psi^t) = \mathcal{N}(0,1), \quad\quad
p(y^t|\psi^t) = \mathcal{N}(\psi^t,\sigma_{y}^2). \label{eq:generative_toy1-2}
\end{equation}
We sample $T=250$ tasks, each with $K=5$ support observations $D^t=\{y_k^t\}_{k=1}^K$, and $M=15$ query observations $\tilde{D}^t=\{\tilde{y}_m^t\}_{m=1}^M$.
We use an  inference network $q_{\phi}(\psi|D^t)=\mathcal{N}(\mu_{q},\sigma_{q}^2)$, where
\bea
\begin{bmatrix} \mu_q  \\ \log \sigma^2_q\end{bmatrix}  =
W \sum_{k=1}^K y_k^t + \vect{b},
\eea
with trainable parameters $W$ and $\vect{b}$.
The inference network is used to define the predictive distribution
\bea
p(\tilde{D}^t|D^t) = \int p(\tilde{D}^t|\psi)q_\phi(\psi|D^t)\;\textrm{d} \psi.
\label{eq:post_pred}
\eea
Since the prior is conjugate to the Gaussian likelihood $p(y^t|\psi^t)$ in \Eq{generative_toy1-2}, we can analytically compute the marginal $p(\tilde{D}^t|D^t)$ in \Eq{post_pred} and the true posterior $p(\psi|D^t)$, which are both Gaussian. 

We train the inference network by optimizing \Eq{post_pred} in the following three ways.
\begin{enumerate}[wide, labelindent=0pt]
    \item {\bf Exact marginal log-likelihood.} For $T$ tasks, with $M$ query samples each, we obtain
    \begin{equation}
       \label{eq:synthetic_loss_analytical}
       \mathcal{L}(\phi)=-\frac{1}{MT}\sum_{t=1}^{T}\sum_{m=1}^{M}\log\mathcal{N}(y_m^t;\mu_q(D^t),\sigma_{q}^2(D^t)+\sigma_y^2).
    \end{equation}

    \item {\bf Monte Carlo estimation.} 
    Using   $L$ samples $\psi_l^t\sim q_{\phi}(\psi|D^t)$ we obtain
    \begin{equation}
       \label{eq:synthetic_loss_MC}
       \mathcal{L}(\phi)=-\frac{1}{MT}\sum_{t=1}^{T}\sum_{m=1}^{M}\log\frac{1}{L}\sum_{l=1}^L\mathcal{N}(y_m^t;\psi_l^t,\sigma_y^2).
    \end{equation}

    \item {\bf Variational inference.} 
    We use the inference network, with a second set of parameters $\phi'$, as variational posterior given both $\tilde{D}^t$ and $D^t$. 
    Using $L$ samples $\psi_l^t\sim q_{\phi'}(\psi|\tilde{D}^t,D^t)$, 
    we obtain
    \begin{equation}
       \label{eq:synthetic_loss_CVAE}
       \begin{split}
       \mathcal{L}(\phi) = &-\frac{1}{T}\sum_{t=1}^{T}\left[\sum_{m=1}^{M}\frac{1}{L}\sum_{l=1}^L\log\mathcal{N}(y_m^t;\psi_l^t,\sigma_y^2)\right. \\
       &\left.-\mathcal{D}_{\textrm{KL}}(q_{\phi'}(\psi|\tilde{D}^t,D^t)||q_{\phi}(\psi|D^t))\vphantom{\sum_{m=1}^{M}}\right].
       \end{split}
    \end{equation}
\end{enumerate}
We trained with these three approaches for $\sigma_y \in \{0.1, 0.5, 1.0\}$.
For Monte Carlo and variational methods, we used the re-parameterization trick to differentiate through sampling $\psi$ \citep{kingma14iclr,rezende14icml}.
We evaluate the quality of the trained inference network by sampling data $D^t$ for a new task from the data generating process \Eq{generative_toy1-2}.
For new data, we compare the true posterior $p(\psi|D^t)$ with the distribution $q_\phi(\psi|D^t)$ produced by the trained inference network.

Results in \fig{synthetic_relative_sigma} show that both the analytic and variational approaches recover true posterior very well, including variational training with a single sample. 
Monte Carlo training, on the other hand, requires the use of significantly larger sets of samples to produce results comparable to other two approaches.
Optimization with a small number of samples leads to significant underestimation of the target variance.
This makes the Monte Carlo training approach either computationally expensive, or inaccurate in modeling the uncertainty in the latent variable.

\subsection{Experimental Setup for Image Classification}
\label{sec:samovar_implementation}

{\bf MiniImageNet} \cite{vinyals2016matching} consists of 100 classes selected from ILSVRC-12 \cite{russakovsky15ijcv}. We follow the split from \citet{ravi17iclr} with 64 meta-train, 16 meta-validation and 20 meta-test classes, and 600 images in each class. 
Following \citet{oreshkin18nips}, we use a central square crop, and resize it to $84\!\times\!84$ pixels.

{\bf FC100} \cite{oreshkin18nips} was derived from CIFAR-100 \cite{krizhevsky2009learning}, which consists of 100 classes, with 600 32$\times$32 images per class. 
All classes are grouped into 20 superclasses. The data is split by superclass to minimize the information overlap. There are 60 meta-train classes from 12 superclasses, 20 meta-validation, and meta-test classes, each from four corresponding superclasses. 

{\bf CIFAR-FS} \cite{bertinetto2018metalearning} is another meta-learning dataset derived from CIFAR-100. It was created by a random split into 64 meta-train, 16 meta-validation and 20 meta-test classes. For each class, there are 600 images of size 32$\times$32.

\mypar{Network architectures and training specifications} For a fair comparison with VERSA \cite{gordon2018metalearning}, we follow the same experimental setup, including the network architectures, optimization procedure, and episode sampling. In particular, we use the shallow {\bf CONV-5} feature extractor. In other experiments we use {\bf ResNet-12} backbone feature extractor \cite{oreshkin18nips,mishra2017asn}. The cosine classifier is scaled by setting $\alpha$ to $25$ when data augmentation is not used, and $50$ otherwise.  The hyperparameters were chosen through cross-validation.
The TEN network used for task conditioning is the same as in \citet{oreshkin18nips}. 
The main and auxiliary tasks are trained concurrently: in episode $t$ out of $T$, the auxiliary task is sampled with probability $\rho = 0.9^{\lfloor12t/T\rfloor}$. 
The choice of $\beta$, as well as other details about the architecture and training procedure can be found in the supplementary material. We provide implementaion of our method at: \url{https://github.com/katafeya/samovar}.

Unless explicitly mentioned, we do not use data augmentation. 
In cases where we do use augmentation, it is performed with random horizontal flips, random crops, and color jitter (brightness, contrast and saturation).

\mypar{Evaluation} We evaluate classification accuracy by randomly sampling 5,000 episodes, and 15 queries per class in each test episode.
We also report 95\% confidence intervals computed over these 5,000 tasks. 
We draw $d=1,000$ samples for each class $n$ from the corresponding prior to make a prediction, and average the resulting probabilities for the final classification. 

\begin{table*}[t]
\renewcommand\thetable{2}
\caption{Accuracy and 95\% confidence intervals of TADAM and SAMOVAR on the 5-way classification task on miniImageNet. The first  columns indicate the use of: cosine scaling ($\alpha$), auxiliary co-training (AT), and task embedding network (TEN). 
}
\label{tab:tadam_comparison}
\vspace*{\baselineskip}
\centering
\begin{small}
\begin{sc}
\begin{tabular}{ccccccc}
\toprule
 & & & \multicolumn{2}{c}{5-shot} & \multicolumn{2}{c}{1-shot} \\
$\alpha$ & AT & TEN & TADAM & SAMOVAR & TADAM & SAMOVAR \\
\midrule
 &  &  & 73.5 $\pm$ 0.2 & 75.3 $\pm$ 0.2 & 58.2 $\pm$ 0.3 & 59.3 $\pm$ 0.3 \\
$\checkmark$ &  &  & 74.9 $\pm$ 0.2 & 76.9 $\pm$ 0.2 & 57.4 $\pm$ 0.3 & 58.2 $\pm$ 0.3 \\
 & $\checkmark$ &  & 74.6 $\pm$ 0.2 & 76.4 $\pm$ 0.2 & 58.7 $\pm$ 0.3 & 59.8 $\pm$ 0.3 \\
 &  & $\checkmark$ & 72.9 $\pm$ 0.2 & 74.9 $\pm$ 0.2 & 58.2 $\pm$ 0.3 & 58.8 $\pm$ 0.3 \\
$\checkmark$ & $\checkmark$ &  & 75.7 $\pm$ 0.2 & 77.2 $\pm$ 0.2 & 57.3 $\pm$ 0.3 & 60.4 $\pm$ 0.3 \\
 & $\checkmark$ & $\checkmark$ & 74.1 $\pm$ 0.2 & 77.3 $\pm$ 0.2 & 57.5 $\pm$ 0.3 & 59.5 $\pm$ 0.3 \\
 $\checkmark$ &  & $\checkmark$ & 74.9 $\pm$ 0.2 & 76.8 $\pm$ 0.2 & 57.3 $\pm$ 0.3 & 58.5 $\pm$ 0.3 \\
$\checkmark$ & $\checkmark$ & $\checkmark$ & 75.9 $\pm$ 0.2 & 77.5 $\pm$ 0.2 & 57.6 $\pm$ 0.3 & 60.7 $\pm$ 0.3 \\
\bottomrule    
\end{tabular}
\end{sc}
\end{small}
\end{table*}

\begin{table}
\renewcommand\thetable{1}
\caption{Accuracy and 95\% confidence intervals of VERSA and SAMOVAR on the 5-way classification task on miniImageNet. Both approaches train the same meta-learning model.
}
\label{tab:versa_comparison}
\vspace*{\baselineskip}
\centering
\begin{small}
\begin{sc}
\resizebox{\columnwidth}{!}{\begin{tabular}{lcc}
\toprule
               & 5-shot           & 1-shot           \\ 
\midrule
VERSA (our implem.) & 68.0 $\pm$ 0.2 & 52.5 $\pm$ 0.3 \\
SAMOVAR-base   & 69.8 $\pm$ 0.2 & 52.4 $\pm$ 0.3 \\
SAMOVAR-base (separate)   & 66.6 $\pm$ 0.2 & 50.8 $\pm$ 0.3 \\
\bottomrule    
\end{tabular}}
\end{sc}
\end{small}
\end{table}

\subsection{Few-Shot Image Classification Results}
\label{sec:results}

\mypar{Comparison with VERSA}
\label{sec:comparison_versa}
In our first experiment, we compare SAMOVAR-base with VERSA \cite{gordon2018metalearning}. Both use the same model, but differ only in their training procedure.
We used the code provided by \citet{gordon2018metalearning} to implement both approaches, making one important change: we avoid compression artefacts by storing image crops in PNG rather than JPG format, which improves results noticeably.

In \tab{versa_comparison} we report the accuracy on miniImageNet for both the models.  
In the 1-shot setup, both the approaches lead to similar results, while SAMOVAR yields considerably better performance in the 5-shot setup. When training VERSA we keep track of the largest variance predicted for model parameters, and observe that it quickly deteriorates from the beginning of training.
We do not observe this collapse in SAMOVAR. This is consistent with the results obtained on synthetic data. More details about distribution collapse in VERSA are presented in the supplementary material.

To evaluate the effect of sharing the inference network between prior and posterior, we run SAMOVAR-base with separate neural networks for prior and posterior, and with the reduced number of hidden units to even out the total number of parameters.
From the results in the last two lines of \tab{versa_comparison}, it can be seen that for both 1-shot and 5-shot classification sharing the inference network has a positive impact on the performance.

\begin{figure}[!ht]
    \vspace*{\baselineskip}
    \begin{center}
    \begin{subfigure}[t]{0.49\columnwidth}
    \includegraphics[width=\textwidth]{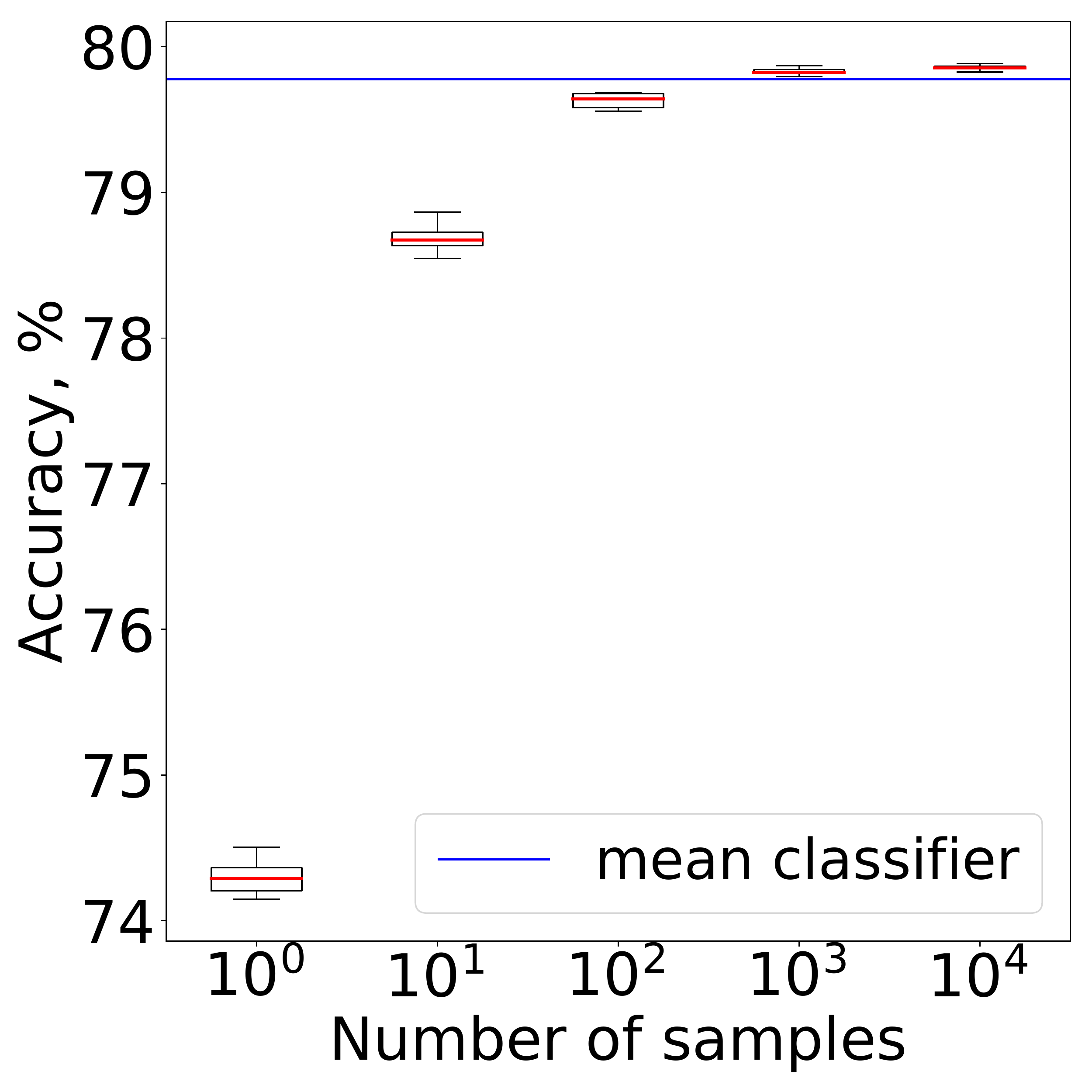}
    \caption{5-shot.}
    \label{fig:accuracy_from_samples_5}
    \end{subfigure}
    \begin{subfigure}[t]{0.49\columnwidth}
    \includegraphics[width=\textwidth]{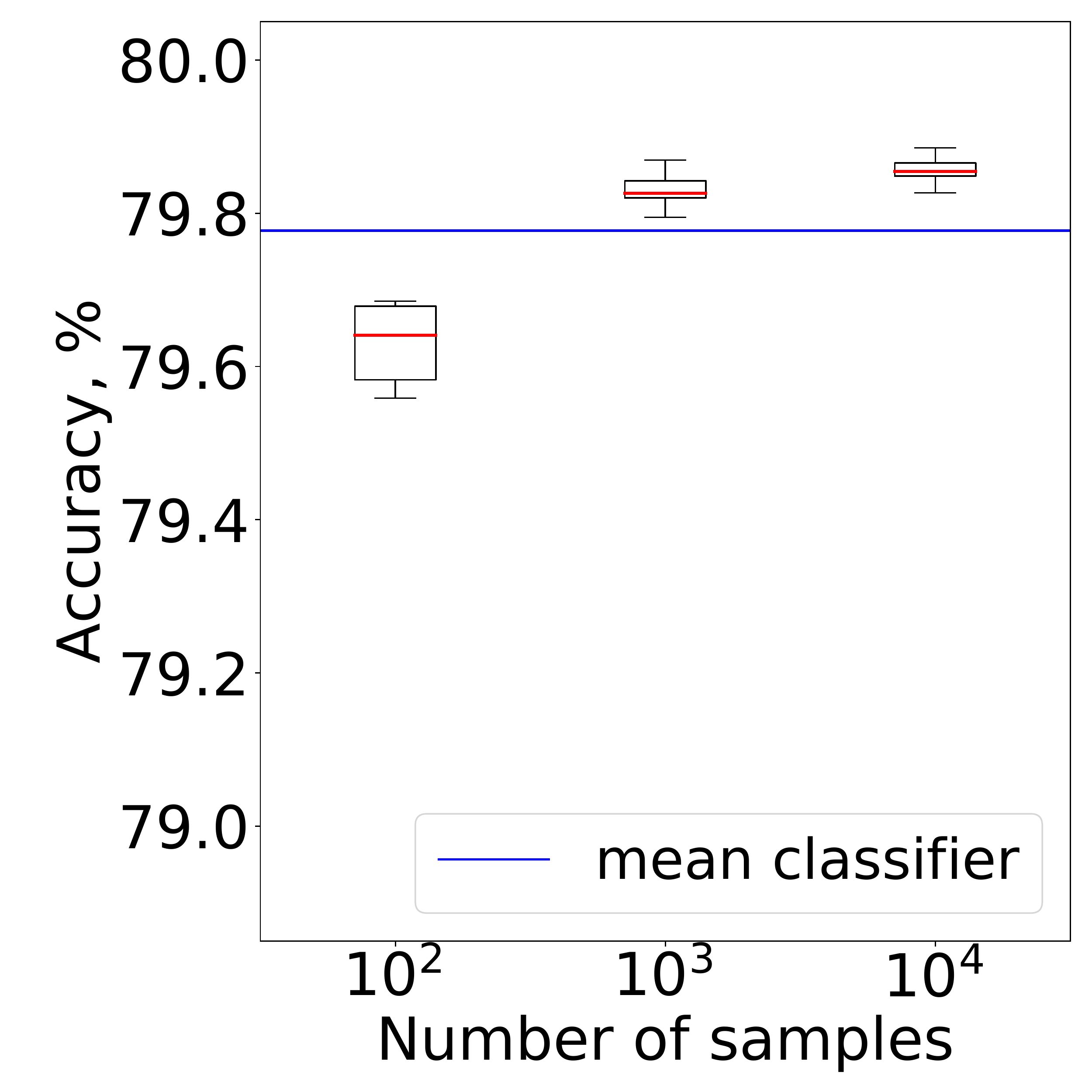}
    \caption{5-shot, zoomed.}
    \label{fig:accuracy_from_samples_5_zoomed}
    \end{subfigure}
    \begin{subfigure}[t]{0.49\columnwidth}
    \includegraphics[width=\textwidth]{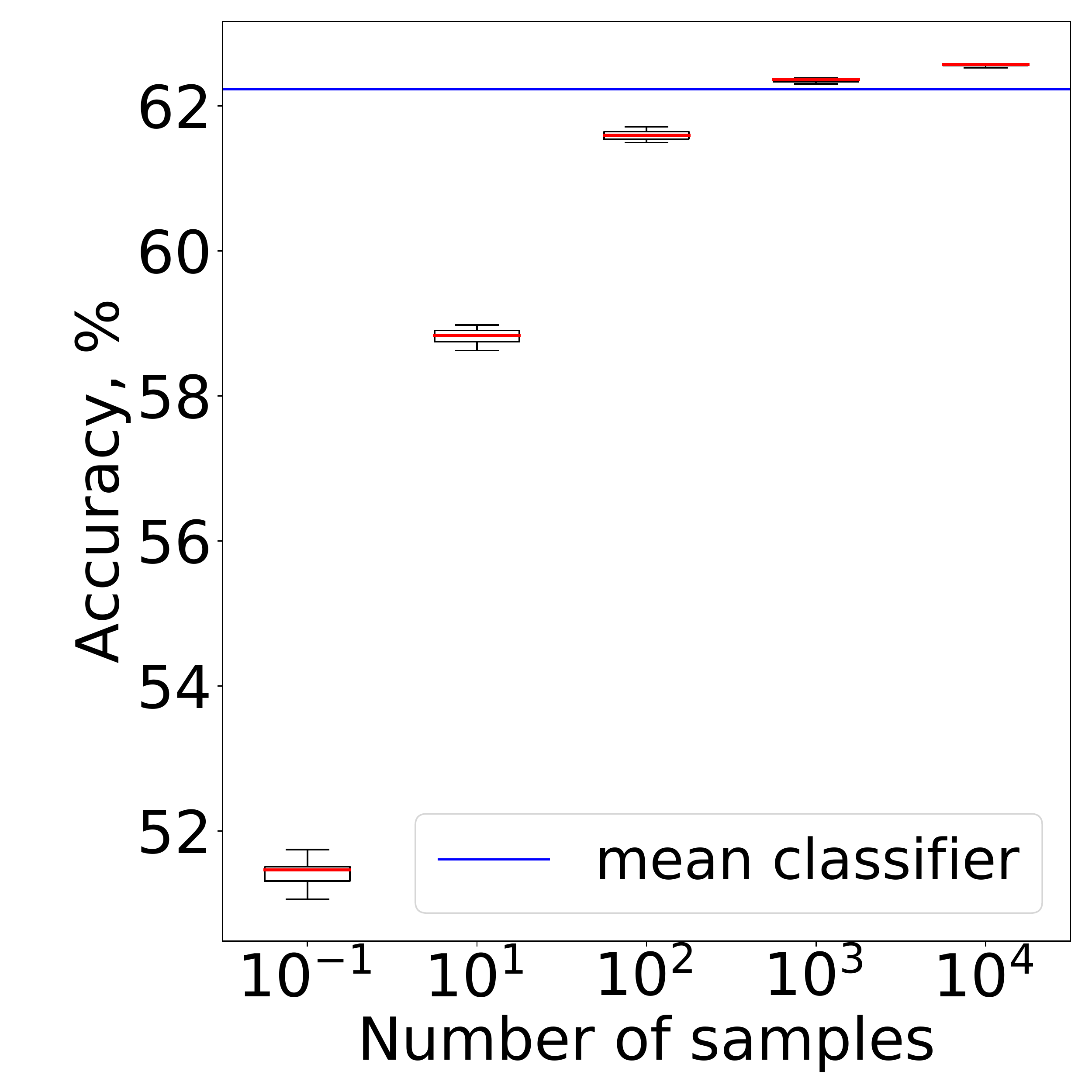}
    \caption{1-shot.}
    \label{fig:accuracy_from_samples_1}
    \end{subfigure}
    \begin{subfigure}[t]{0.49\columnwidth}
    \includegraphics[width=\textwidth]{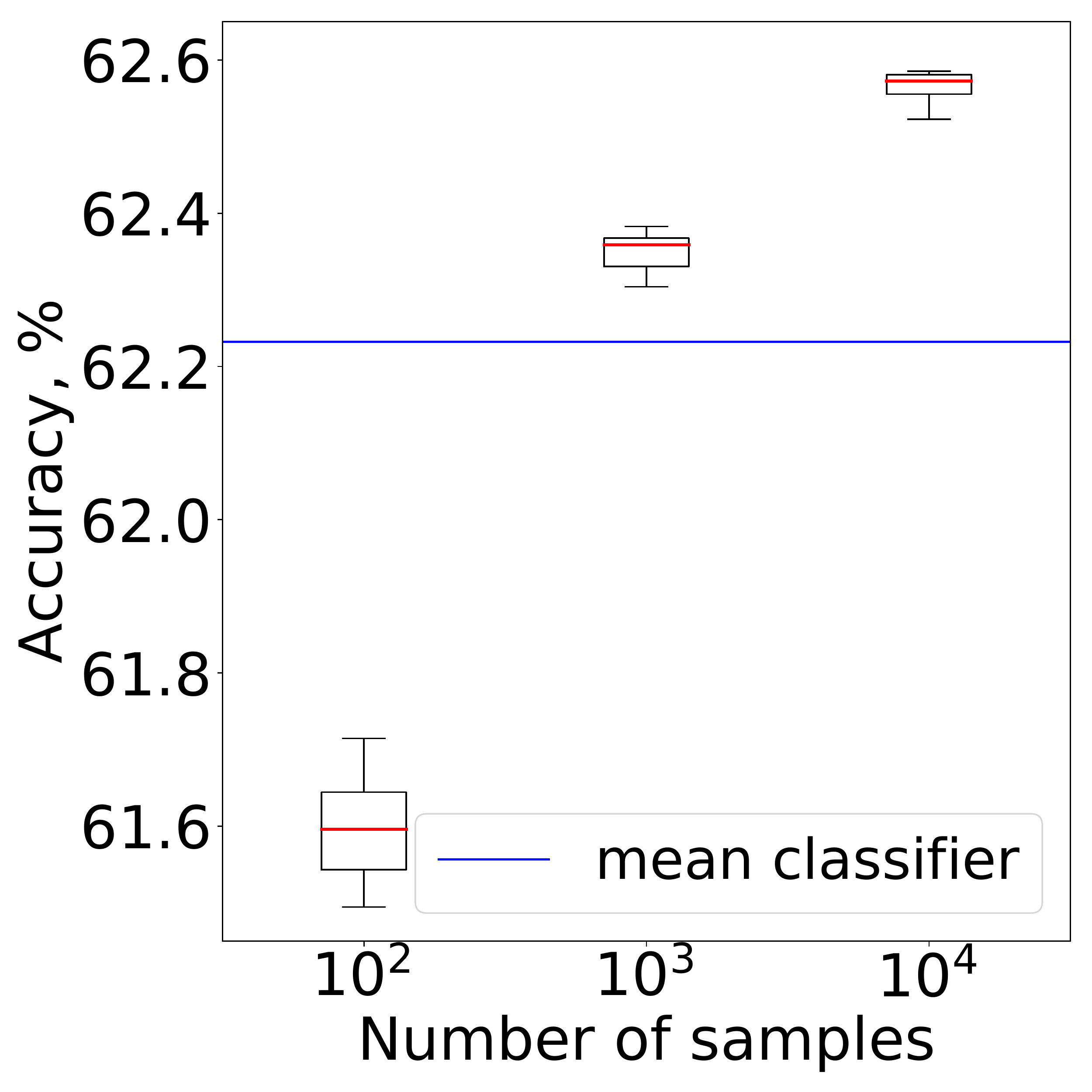}
    \caption{1-shot, zoomed.}
    \label{fig:accuracy_from_samples_1_zoomed}
    \end{subfigure}
    \end{center}
    \caption{Accuracy on miniImageNet as a function of the number of samples drawn from the learned prior over the classifier weights, compared to using the mean of the distribution. 
    }
    \label{fig:accuracy_from_samples}
\end{figure}

\begin{table*}
\caption{Accuracy and 95\% confidence intervals of state-of-the-art models on the 5-way task on miniImageNet. Versions of the models that use additional data during training are not included. Exception is made only if this is the sole result provided by the authors.
$\ast$: Results obtained with data augmentation.
$\dagger$: Transductive methods.
$\circ$: Validation set is included into training.
$\bigtriangleup$: Based on a 1.25$\times$wider ResNet-12 architecture.
}
\label{tab:model_comparison_full}
\vspace*{\baselineskip}
\centering
\begin{small}
\begin{sc}
\resizebox{\textwidth}{!}{
\begin{tabular}{lllll}
    \toprule
    Method & Features  & 5-shot & 1-shot & Test protocol\\
    \midrule
    Matching Nets\citep{vinyals2016matching} & CONV-4 & 60.0 & 46.6 & \\
    Meta LSTM\citep{ravi17iclr} & CONV-4 & 60.6 $\pm$ 0.7 & 43.4 $\pm$ 0.8 & 600 ep. / 5$\times$15 \\
    MAML \citep{finn17icml} & CONV-4 & 63.1 $\pm$ 0.9 & 48.7 $\pm$ 1.8 & 600 ep. / 5 $\times$ shot \\
    RelationNet \citep{sung2018learning} & CONV-4 & 65.3 $\pm$ 0.7 & 50.4 $\pm$ 0.8 & 600 ep. / 5 $\times$ 15 \\
    Prototypical Nets \citep{snell2017prototypical} & CONV-4 & 65.8 $\pm$ 0.7 & 46.6 $\pm$ 0.8 & 600 ep. / 5 $\times$ 15 \\
    VERSA \citep{gordon2018metalearning} & CONV-5 & 67.4 $\pm$ 0.9 & 53.4 $\pm$ 1.8 & 600 ep. / 5 $\times$ shot \\
    TPN \citep{liu2019fewTPN} & CONV-4$^{\dagger}$ & 69.9 & 55.5 & 2000 ep. / 5 $\times$ 15 \\
    SIB\citep{hue2020empirical} & CONV-$4^{\dagger}$ & 70.7 $\pm$ 0.4 & 58.0 $\pm$ 0.6 & 2000 ep. / 5 $\times$ 15 \\
    \citet{gidaris2019boosting} & CONV-4 & 71.9 $\pm$ 0.3 & 54.8 $\pm$ 0.4 & 2000 ep. / 5 $\times$ 15 \\
    SAMOVAR-BASE (ours) & CONV-5 & 69.8 $\pm$ 0.2 & 52.4 $\pm$ 0.3 & 5000 ep. / 5 $\times$ 15 \\
    \midrule
    \citet{qiao2018few} & WRN-28-10 & 73.7 $\pm$ 0.2 & 59.6 $\pm$ 0.4 & 1000 ep. / 5 $\times$ 15 \\
    MTL HT \citep{sun2019mtl} & ResNet-12 & 75.5 $\pm$ 0.8 & 61.2 $\pm$ 1.8 & 600 ep. / 5 $\times$ shot \\
    TADAM \citep{oreshkin18nips} & ResNet-12 & 76.7 $\pm$ 0.3 & 58.5 $\pm$ 0.3 & 5000 ep. / 100 \\
    LEO \citep{rusu19iclr} & WRN-28-10$^{\ast \circ}$ & 77.6 $\pm$ 0.1 & 61.8 $\pm$ 0.1 & 10000 ep. / 5 $\times$ 15 \\
    Fine-tuning \citep{dhillon2020baseline} & WRN-28-10$^{\ast}$ & 78.2 $\pm$ 0.5 & 57.7 $\pm$ 0.6 & 1000 ep. / 5 $\times$ 15 \\
    Transductive Fine-tuning \citep{dhillon2020baseline} & WRN-28-10$^{\ast \dagger}$ & 78.4 $\pm$ 0.5 & 65.7 $\pm$ 0.7 & 1000 ep. / 5 $\times$ 15 \\
    MetaOptNet-SVM \citep{lee2019meta} & ResNet-12$^{\ast\bigtriangleup}$ & 78.6 $\pm$ 0.5 & 62.6 $\pm$ 0.6 & 2000 ep. / 5 $\times$ 15 \\
    SIB \citep{hue2020empirical} & WRN-28-10$^{\ast \dagger}$ & 79.2 $\pm$ 0.4 & 70.0 $\pm$ 0.6 & 2000 ep. / 5 $\times$ 15 \\
    \citet{gidaris2019boosting} & WRN-28-10$^{\ast}$ & 79.9 $\pm$ 0.3 & 62.9 $\pm$ 0.5 & 2000 ep. / 5 $\times$ 15 \\
    CTM \citep{li2019ctm} & ResNet-18$^{\ast \dagger}$ & 80.5 $\pm$ 0.1 & 64.1 $\pm$ 0.8 & 600 ep. / 5 $\times$ 15 \\
    \citet{dvornik19iccv} & WRN-28-10$^{\ast}$ & 80.6 $\pm$ 0.4 & 63.1 $\pm$ 0.6 & 1000 ep. / 5 $\times$ 15 \\
    SAMOVAR-SC-AT-TEN (ours) & ResNet-12 & 77.5 $\pm$ 0.2 &  60.7 $\pm$ 0.3 & 5000 ep. / 100 \\
    SAMOVAR-SC-AT-TEN (ours) &  ResNet-12$^{\ast}$ & 79.5 $\pm$ 0.2 & 63.3 $\pm$ 0.3 & 5000 ep. / 5 $\times$ 15 \\
    \bottomrule
\end{tabular}
}

\end{sc}
\end{small}
\end{table*}

\mypar{Comparison with TADAM}
In our second experiment, we use SAMOVAR in combination with the architecture of TADAM \cite{oreshkin18nips}. 
To fit our framework, we replace the prototype classifier of TADAM with a linear classifier with latent weights.
We compare TADAM and SAMOVAR with metric scaling ($\alpha$), auxiliary co-training (AT) and the task embedding network (TEN) included or not.
When the metric is not scaled, we use SAMOVAR-base with the linear classifier, otherwise we use SAMOVAR-SC with the scaled cosine classifier. 
For this ablative study we fix the random seed to generate the same series of meta train, meta validation and meta test tasks for both models, and for all configurations.
The results in \tab{tadam_comparison} show that SAMOVAR provides a consistent improvement over TADAM across all the tested ablations of the TADAM architecture.

\mypar{Effect of sampling classifier weights}
To assess the effect of the stochasticity of the model, we evaluate the prediction accuracy obtained with the mean of the distribution on classifier weights, and approximating the predictive distribution of \Eq{predictive_distribution} with a varying number of samples of the classifier weights.
For both the 5-shot and 1-shot setups, we fix the random seed and evaluate SAMOVAR-SC-AT-TEN on the same 1,000 random 5-way tasks. 
We compute accuracy 10 times for each number of samples. 

Results of these experiments for 5-shot and 1-shot tasks are shown in  \fig{accuracy_from_samples}. 
It can be seen that for both setups the mean classification accuracy is positively correlated with the number of samples. 
This is expected as a larger sample size corresponds to a better estimation of the predictive posterior distribution. 
The dispersion of accuracy for a fixed $n$ is slightly bigger for the 1-shot setup compared to the 5-shot setup, and in both cases it decreases as we use more samples. 
This difference is also expected, as the 1-shot task is much harder than the 5-shot task, so the model retains more uncertainty in the inference in the former case. 
The results also show that the predicted classifier mean demonstrates good results on both classification tasks, and it can be used instead of classifier samples in cases where computational budget is critical. 
At the same time we can see that sampling of a large number of classifiers leads to a better performance compared to the classifier mean.
While on the 5-shot setup the gain from  classifier sampling over using the mean is small, around 0.1\% with 10K samples, on the 1-shot setup the model benefits more from the stochasticity yielding additional 0.4\% accuracy with 10K samples.

\begin{table*}
\caption{Accuracy and 95\% confidence intervals of state-of-the-art models on the 5-way task on FC100. Versions of the models that use additional data during training are not included.
$\ast$: Results obtained with data augmentation.
$\square$: Results from \citet{lee2019meta}.
$\dagger$: Transductive methods.
$\bigtriangleup$: Based on a 1.25$\times$wider ResNet-12 architecture.
}
\label{tab:fc100_model_comparison}
\vspace*{\baselineskip}
\centering
\begin{small}
\begin{sc}
\resizebox{\textwidth}{!}{
\begin{tabular}{lllll}
    \toprule
    Method & Features  & 5-shot & 1-shot & Test protocol\\
    \midrule
    Prototypical Nets \citep{snell2017prototypical} & ResNet-12$^{\ast \square\bigtriangleup}$ & 52.5 $\pm$ 0.6 & 37.5 $\pm$ 0.6 & 2000 ep. / 5 $\times$ 15 \\
    TADAM \citep{oreshkin18nips} & ResNet-12 & 56.1 $\pm$ 0.4 & 40.1 $\pm$ 0.4 & 5000 ep. / 100 \\
    MetaOptNet-SVM \citep{lee2019meta} & ResNet-12$^{\ast\bigtriangleup}$ & 55.5 $\pm$ 0.6 & 41.1 $\pm$ 0.6 & 2000 ep. / 5 $\times$ 15 \\
    Fine-tuning \citep{dhillon2020baseline} & WRN-28-10$^{\ast}$ & 57.2 $\pm$ 0.6 & 38.3 $\pm$ 0.5 & 1000 ep. / 5 $\times$ 15 \\
    Transductive Fine-tuning \citep{dhillon2020baseline} & WRN-28-10$^{\ast \dagger}$ & 57.6 $\pm$ 0.6 & 43.2 $\pm$ 0.6 & 1000 ep. / 5 $\times$ 15 \\
    MTL HT \citep{sun2019mtl} & ResNet-12$^{\ast}$ & 57.6 $\pm$ 0.9 & 45.1 $\pm$ 1.8 & 600 ep. / 5 $\times$ shot \\
    \midrule
    SAMOVAR-SC-AT-TEN (ours) &  ResNet-12$^{\ast}$ & 57.9 $\pm$ 0.3 & 42.1 $\pm$ 0.3 & 5000 ep. / 5 $\times$ 15 \\
    \bottomrule
\end{tabular}
}
\end{sc}
\end{small}
\end{table*}

\begin{table*}
\caption{Accuracy and 95\% confidence intervals of state-of-the-art models on the 5-way task on CIFAR-FS. Versions of the models that use additional data during training are not included. All models use data augmentation.
$\square$: Results from \citet{lee2019meta}.
$\dagger$: Transductive methods.
$\bigtriangleup$: Based on a 1.25$\times$wider ResNet-12 architecture.
}
\label{tab:cifar_fs_model_comparison}
\vspace*{\baselineskip}
\centering
\begin{small}
\begin{sc}
\resizebox{\textwidth}{!}{
\begin{tabular}{lllll}
    \toprule
    Method & Features  & 5-shot & 1-shot & Test protocol\\
    \midrule
    Prototypical Nets \citep{snell2017prototypical} & ResNet-12$^{ \square\bigtriangleup}$ & 83.5 $\pm$ 0.5 & 72.2 $\pm$ 0.7 & 2000 ep. / 5 $\times$ 15 \\
    MetaOptNet-SVM \citep{lee2019meta} & ResNet-12$^{\bigtriangleup}$ & 84.2 $\pm$ 0.5 & 72.0 $\pm$ 0.7 & 2000 ep. / 5 $\times$ 15 \\
    Fine-tuning \citep{dhillon2020baseline} & WRN-28-10 & 86.1 $\pm$ 0.5 & 68.7 $\pm$ 0.7 & 1000 ep. / 5 $\times$ 15 \\
    Transductive Fine-tuning \citep{dhillon2020baseline} & WRN-28-10$^{\dagger}$ & 85.8 $\pm$ 0.6 & 76.6 $\pm$ 0.7 & 1000 ep. / 5 $\times$ 15 \\
    SIB \citep{hue2020empirical} & WRN-28-10$^{\dagger}$ & 85.3 $\pm$ 0.4 & 80.0 $\pm$ 0.6 & 2000 ep. / 5 $\times$ 15 \\
    \citet{gidaris2019boosting} & WRN-28-10 & 86.1 $\pm$ 0.2 & 73.6 $\pm$ 0.3 & 2000 ep. / 5 $\times$ 15 \\
    \midrule
    SAMOVAR-SC-AT-TEN (ours) &  ResNet-12 & 85.3 $\pm$ 0.2 & 72.5 $\pm$ 0.3 & 5000 ep. / 5 $\times$ 15 \\
    \bottomrule
\end{tabular}
}
\end{sc}
\end{small}
\end{table*}

\mypar{Comparison to the state of the art}
In \tab{model_comparison_full}, we  compare SAMOVAR to the state of the art on miniImageNet. 
For a fair comparison, we report results with and without  data augmentation. SAMOVAR yields competitive results, notably 
outperforming other approaches using ResNet-12 features.
The only approaches reporting better results explore techniques that are complementary to ours.
Self-supervised co-training was used by 
\citet{gidaris2019boosting}, which can be used as an alternative to  the auxiliary 64-class classification task we used. 
CTM \cite{li2019ctm} is a recent transductive extension to distance-based models, it identifies task-relevant features using inter- and intra-class relations. 
This module  can also be used in conjunction with SAMOVAR, in particular, as an input to the inference network instead of the prototypes.
Finally, knowledge distillation on an ensemble of 20 metric-based classifiers was used by \citet{dvornik19iccv}, which can be used as an alternative  feature extractor  in our work.

In \tab{fc100_model_comparison}, we compare to the state of the art on the  FC100 dataset. We train our model using data augmentation.
SAMOVAR yields the best results on the 5-shot classification task.
Transductive fine-tuning \cite{dhillon2020baseline} reports a higher accuracy for the 1-shot setting, but is not directly comparable due to the transductive nature of their approach.
MTL HT \cite{sun2019mtl} reports the best results (with large 95\% confidence intervals due to the small amount of data used in their evaluation) in the 1-shot setting.
It  samples hard tasks after each meta-batch update by taking its $m$ hardest classes, and makes additional updates of the optimizer on these tasks. 
This is complementary, and can be used in combination with our approach to further improve the results. 

In \tab{cifar_fs_model_comparison}, we compare our model to the state of the art on CIFAR-FS. Data augmentation is used during training. Similar to the aforementioned datasets, SAMOVAR yields competitive results on both tasks. On the 5-shot task, higher accuracy is reported by \citet{dhillon2020baseline} and \citet{gidaris2019boosting}, while transductive SIB \cite{hue2020empirical} is comparable to SAMOVAR. On the 1-shot task, SIB \cite{hue2020empirical}, transductive version by \citet{dhillon2020baseline} and \citet{gidaris2019boosting} report better results. Overall, the observations are consistent with those on miniImageNet.

\section{Conclusion}

We proposed SAMOVAR, a meta-learning model for few-shot image classification that treats classifier weight vectors as latent variables, and uses a shared amortized variational inference network for the prior and variational posterior.
Through experiments on synthetic data and few-shot image classification, we show that our variational approach avoids the severe under-estimation of the variance in the classifier weights observed for training with direct  Monte Carlo approximation~\cite{gordon2018metalearning}. We integrate SAMOVAR with the deterministic TADAM architecture~\citep{oreshkin18nips}, and find that our stochastic formulation leads to significantly improved performance, competitive with the state of the art on the miniImageNet, CIFAR-FS and FC100 datasets.

\section*{Acknowledgements}
We would like to thank the reviewers for their time and constructive comments.
This work was supported in part by the AVENUE project (grant ANR-18-CE23-0011).


\clearpage
\appendix

\section{Network Architectures}
\label{app:architectures}
We learn separate amortized inference networks to predict the mean $\mu$ and log-variance $\ln\sigma^2$ of the latent classification weight vectors $w^t$. 
Both networks have the same architecture, which  depends on the feature extractor that is used. 
The inference networks are shared between the prior and approximate posterior distributions. 

\subsection{CONV-5 Feature Extractor}
The embedding of the image returned by the CONV-5 feature extractor is a 256-dimensional vector. Each of the inference networks for the mean and log variance of the classifier weights $w^t$ consists of three fully connected layers with 256 input and output features, and ELU non-linearity \citep{clevert16iclr} between the layers. There are two additional inference networks that predict the mean and log variance of the classifier biases $b^t$. Both of them consist of two fully connected layers with 256 input and output features followed by ELU non-linearity, and a fully connected layer with 256 input and a single output feature. The design is the same as used by  \citet{gordon2018metalearning} to ensure comparability.

\subsection{ResNet-12 Feature Extractor}
With the ResNet-12 feature extractor, every image is embedded into a 512-dimensional feature vector. Each of the two inference networks consists of three fully connected layers with 512 input and output features, with skip connections and swish-1 non-linearity \citep{ramachandran2017searching} applied before addition in the first two dense layers. 

\section{Training Details for ResNet-12} 
\label{app:training}

For comparison with TADAM \cite{oreshkin18nips} we use the same optimization procedure, number of SGD updates, and weight decay parameters for common parts of the architecture as in the paper. For experiments with data augmentation on miniImageNet we use 40k SGD updates with momentum 0.9, and early stopping based on meta-validation performance. 
We set the initial learning rate to 0.1, and  decrease it by a factor ten after 20k, 25k and 30k updates. 
On FC100 and CIFAR-FS, we use 30k SGD updates with the same momentum and initial learning rate, and the latter is decreased after 15k, 20k and 25k updates.
We clip gradients at 0.1, and set separate weight decay rates for the feature extractor, TEN, fully connected layer in the auxiliary task, and inference networks. For the feature extractor and TEN the weight decay is 0.0005. For the fully connected layer in the auxiliary task the weight decay is 0.00001 on miniImageNet, and 0.0005 on FC100 and CIFAR-FS. In the 1-shot setup, the inference networks are regularized with the weight decay equal to 0.0005, regardless of the dataset. In the 5-shot setup, the weight decay parameter in the inference networks is 0.00001 on miniImageNet, and 0.00005 on FC100 and CIFAR-FS.
We empirically find that the regularization coefficient $\beta = \frac{K}{Nd}$ produces good results, and it can be used as a starting point for further parameter tuning. Here $d$ is the dimensionality of the feature vector $f_{\theta}$, $N$ is the number of classes in the task, and $K$ is the total number of query samples in the task. On CONV-5, we set $\beta$ to 0.0586 for the 5-shot setup, and we multiply it by two for the 1-shot setup. On ResNet-12, we set $\beta$ to 0.0125 for both setups, and we use a value of $\beta$ twice as large for the 1-shot setup without auxiliary co-training.

For the 5-shot setup, mini-batches consist of two episodes, each with 32 query images. For the 1-shot setup, we sample 5 episodes per mini-batch, and 12 query images per episode. In both cases query images are sampled uniformly across classes, without any restriction on the number per class. The auxiliary 64-way classification task is trained with the batch size 64.

\section{Impact of $\beta$-scaling}
\label{app:beta}

\begin{figure}
\vspace{0.4\baselineskip}
\begin{center}
\begin{subfigure}[t]{0.49\columnwidth}
\includegraphics[width=\columnwidth]{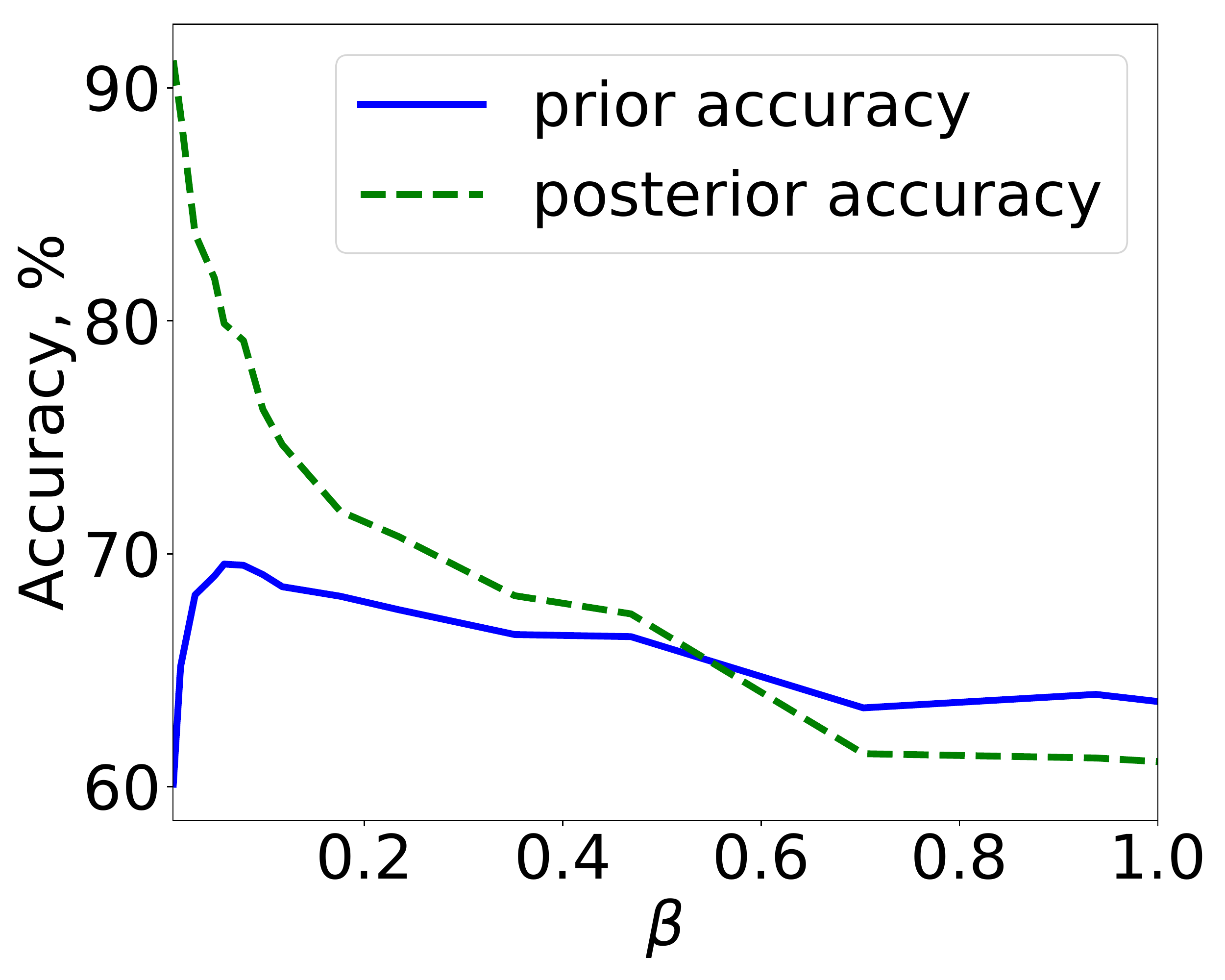}
\caption{5-shot setup}
\end{subfigure}
\begin{subfigure}[t]{0.49\columnwidth}
\includegraphics[width=\columnwidth]{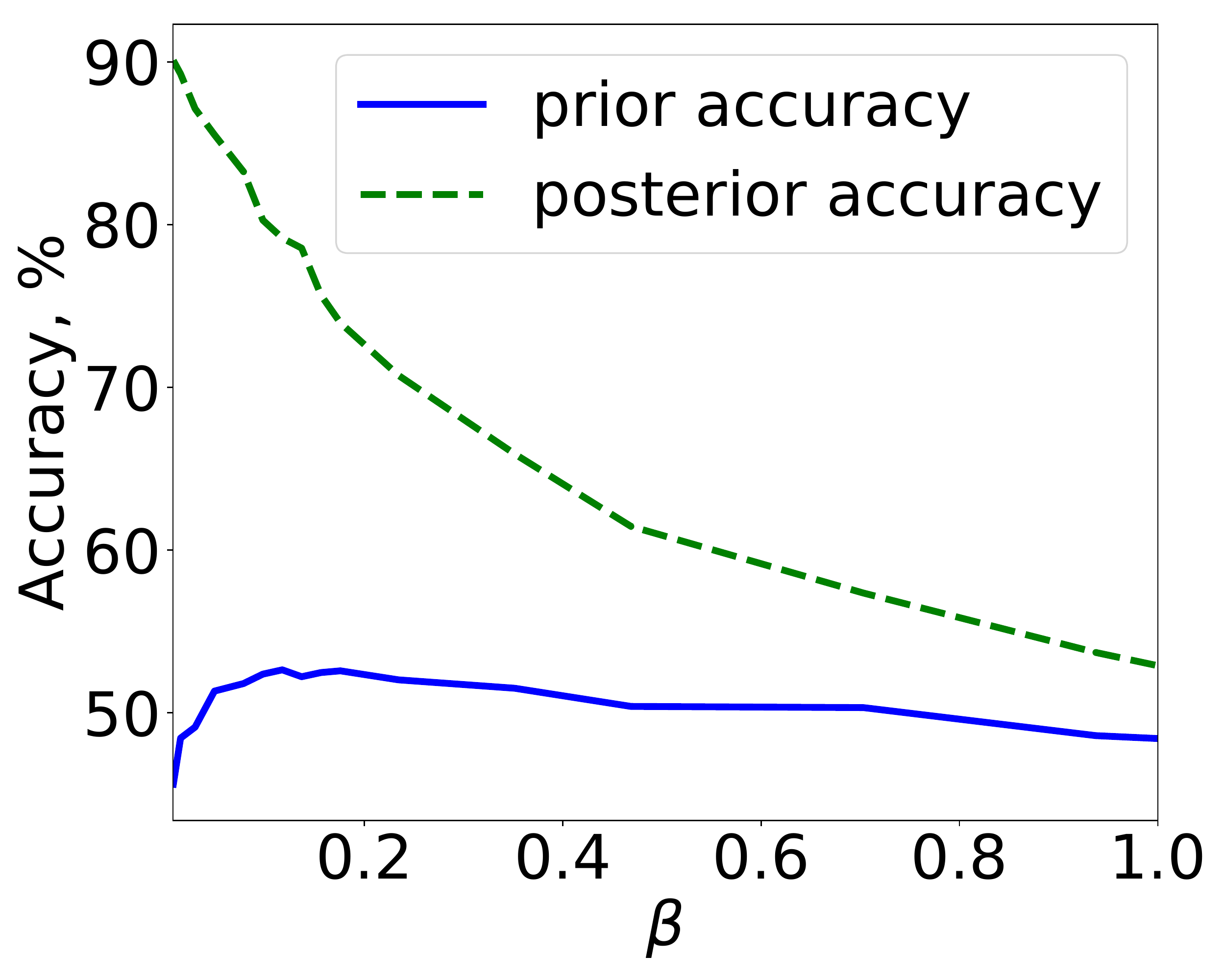}
\caption{1-shot setup}
\end{subfigure}
\end{center}
\caption{Mean accuracy of the SAMOVAR-base classifiers sampled from the prior and posterior as a function of $\beta$. While training, we fix the random seed of the data to generate the same series of miniImageNet tasks. The evaluation is performed over 5000 random tasks.
}
\label{fig:beta_ablation}
\end{figure}

\begin{figure}[!t]
    \vspace*{\baselineskip}
    \centering
    \begin{subfigure}[t]{0.49\columnwidth}
    \includegraphics[width=\textwidth]{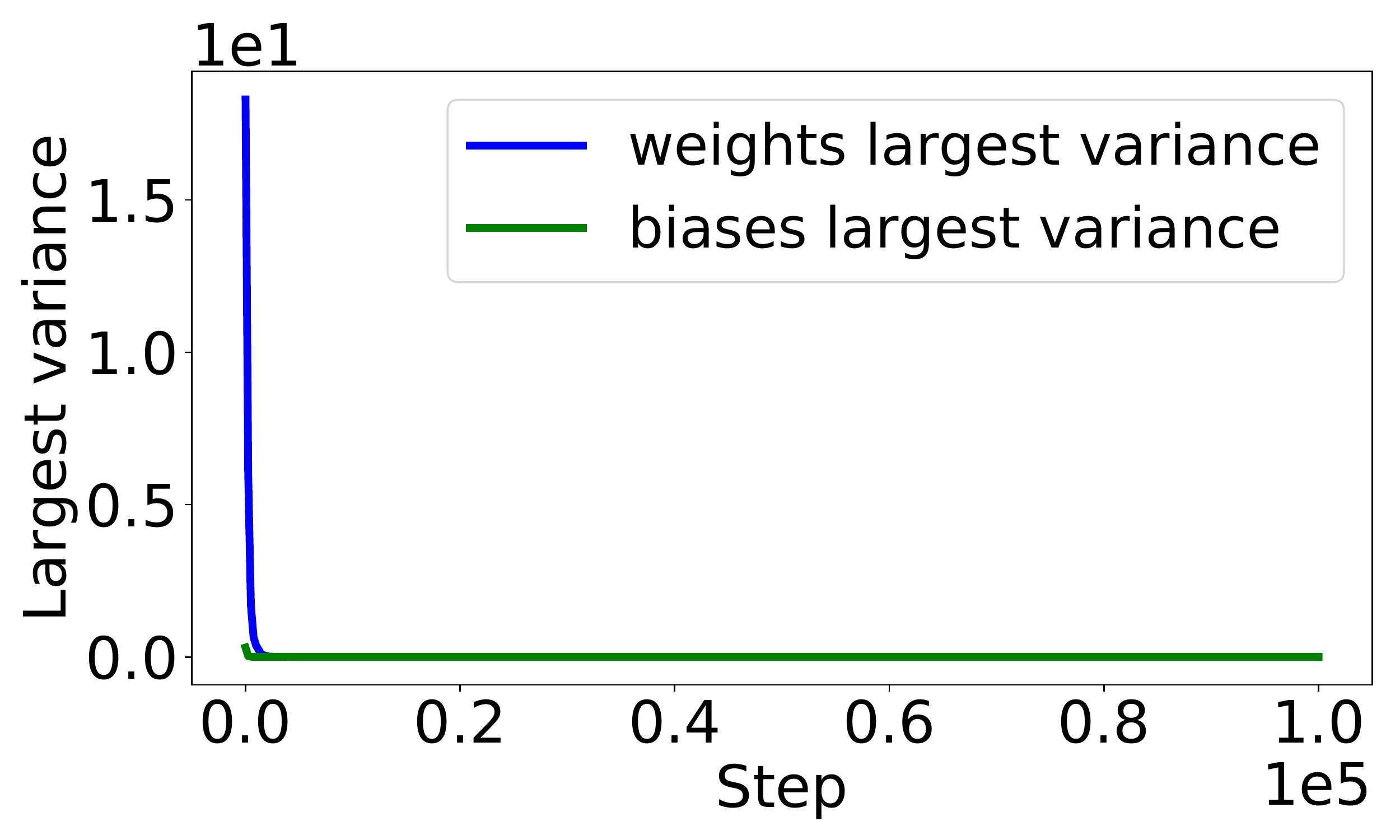}
    \caption{5-shot setup.}
    \label{fig:versa_var_s5}
    \end{subfigure}
    \begin{subfigure}[t]{0.49\columnwidth}
    \includegraphics[width=\textwidth]{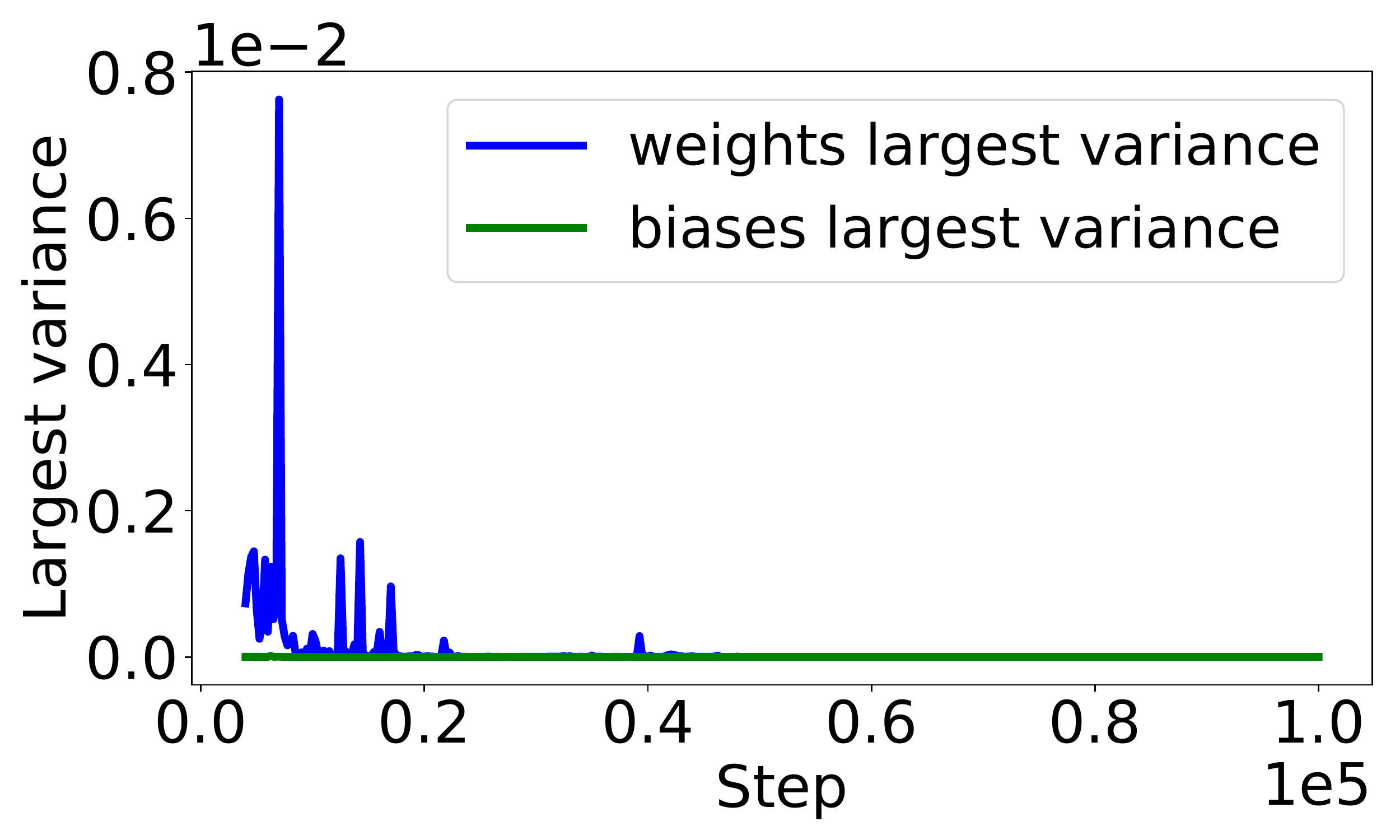}
    \caption{5-shot setup zoomed in.}
    \label{fig:versa_var_s5_zoomed}
    \end{subfigure}
    \begin{subfigure}[t]{0.49\columnwidth}
    \includegraphics[width=\textwidth]{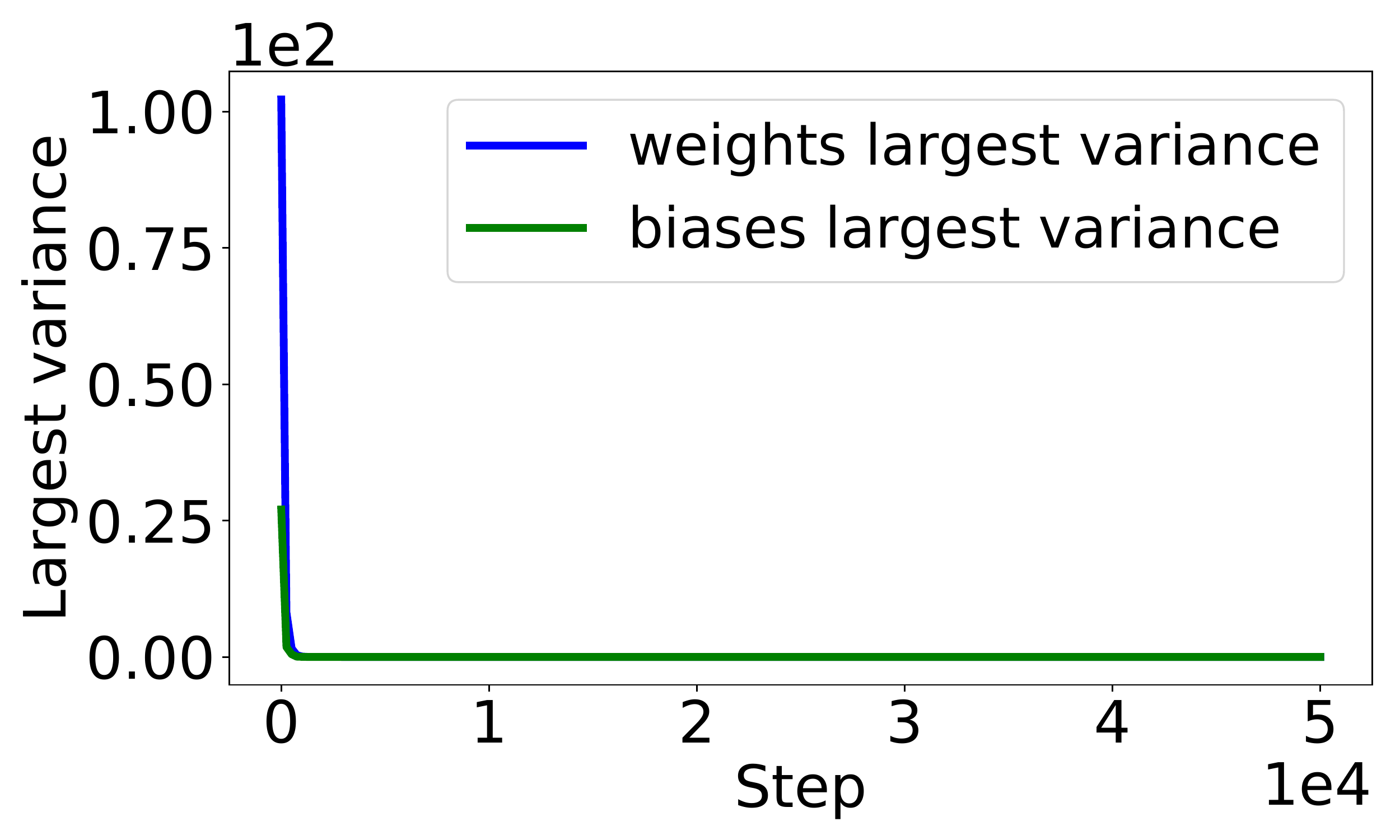}
    \caption{1-shot setup.}
    \label{fig:versa_var_s1}
    \end{subfigure}
    \begin{subfigure}[t]{0.49\columnwidth}
    \includegraphics[width=\textwidth]{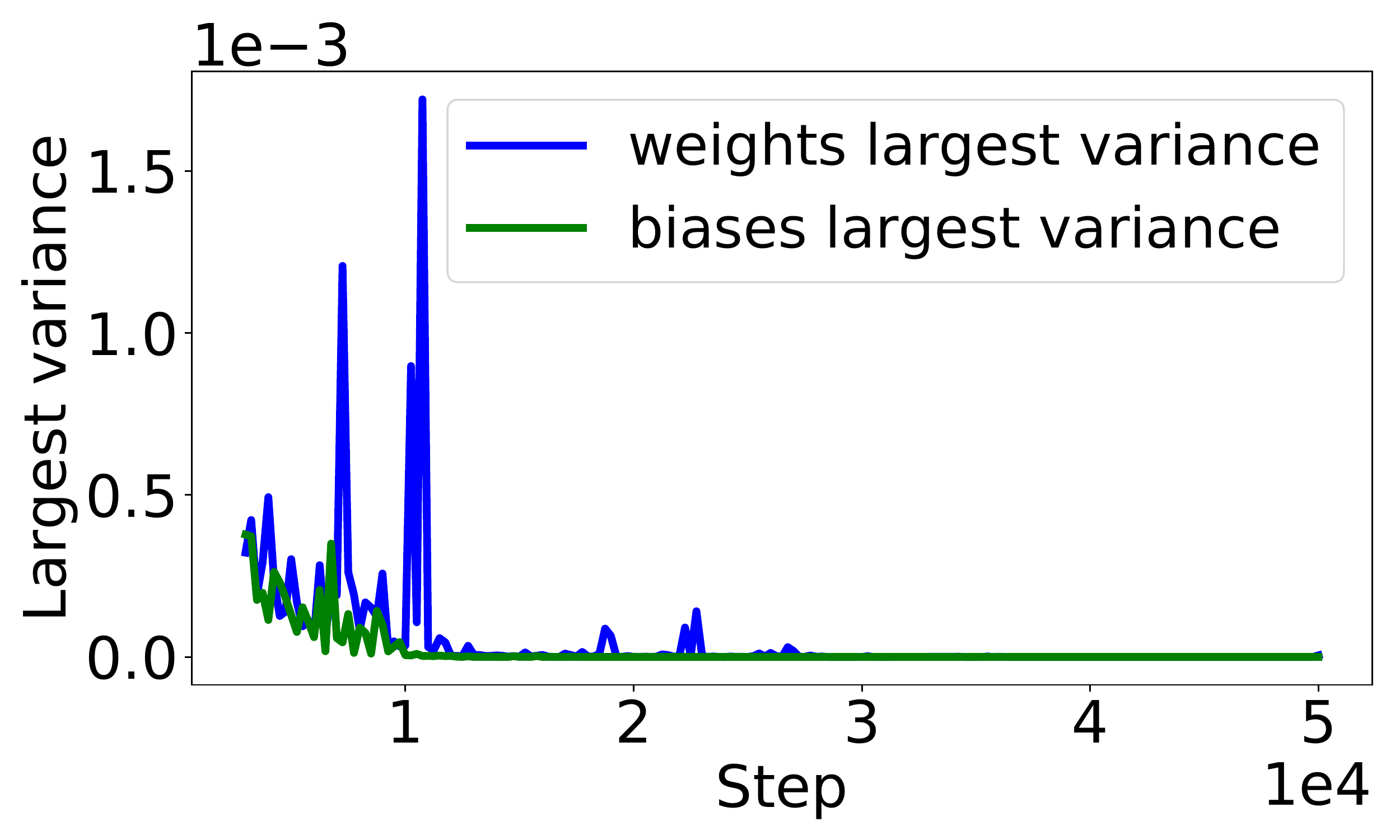}
    \caption{1-shot setup zoomed in.}
    \label{fig:versa_var_s1_zoomed}
    \end{subfigure}
    \caption{Largest variance in VERSA  as a function of the optimization step. Results for optimization steps from \fig{versa_var_s5} and \fig{versa_var_s1} that follow the first encounter of variance below 0.001 are zoomed in \fig{versa_var_s5_zoomed} \fig{versa_var_s1_zoomed} respectively.
    }
    \label{fig:versa_variance_analysis}
\end{figure}

Typically, in autoencoders the dimensionality of the latent space is smaller than of the observed. This is not the case in the meta learning classification task where the output is merely a one-hot-encoded label of the class, while the latent space is of the same size as the output of the feature extractor. In our experiments we observe that the large KL term suppresses the reconstruction term resulting in a weaker performance. In particular, there is a trade off between these parts of the objective function $\hat{\mathcal{L}}(\Theta)$ which can be regulated by $\beta$-scaling of the KL term. \fig{beta_ablation} shows the accuracy of  SAMOVAR-base with CONV-5 feature extractor as a function of $\beta$. Even though in both setups there is a clear maximum, overall, the model is relatively robust to the setting of $\beta$. 
Let's denote the optimum $\beta$ as $\beta_{\text{opt}}$. Then for the 5-shot setup the range at least from $0.83\beta_{\text{opt}}$ to $2\beta_{\text{opt}}$ produces results that are within the 1\% interval from the maximum accuracy at $\beta_{\text{opt}}$. For the 1-shot setup, the same holds true for the range at least from $0.66\beta_{\text{opt}}$ to $2\beta_{\text{opt}}$.

\section{Posterior Collapse in VERSA}
\label{app:versa_collapse}

While training VERSA, every 250 optimization steps we keep track of the largest variance of the weights and biases of the predicted classifier. \fig{versa_variance_analysis} shows how this variance decreases with time. For example, the largest variance of the weights first falls below 0.001 at the step ~4000 in the 5-shot setup, and at the step ~3000 in the 1-shot setup.

\end{document}